\documentclass[manuscript]{acmart}
\usepackage{graphicx}%
\usepackage{multirow}%
\usepackage{amsmath,amsfonts}%
\usepackage{amsthm}%
\usepackage{mathrsfs}%
\usepackage[title]{appendix}%
\usepackage{xcolor}%
\usepackage{textcomp}%
\usepackage{manyfoot}%
\usepackage{booktabs}%
\usepackage{pgfplots}
\usepackage{algorithm}%
\usepackage{algorithmicx}%
\usepackage{algpseudocode}%
\usepackage{listings}%
\usepackage[normalem]{ulem}  
\usepackage{xcolor}   
\usepackage[normalem]{ulem}
\useunder{\uline}{\ul}{}

\usepackage{adjustbox}
\usepackage[normalem]{ulem}
\useunder{\uline}{\ul}{}
\usepackage{booktabs}
\usepackage{array}
\usepackage{siunitx}
\usepackage[normalem]{ulem}
\useunder{\uline}{\ul}{}
\usepackage{natbib}
\usepackage{caption}
\usepackage{booktabs}
\usepackage{siunitx}
\usepackage{graphicx}
\usepackage{tikz}
\usepackage{graphicx}
\theoremstyle{thmstyleone}%
\theoremstyle{thmstyletwo}%

\theoremstyle{thmstylethree}%
\newtheorem{definition}{Definition}%

\AtBeginDocument{%
  }

\setcopyright{acmlicensed}
\copyrightyear{2018}
\acmYear{2018}
\acmDOI{XXXXXXX.XXXXXXX}

\acmJournal{JDS}
\acmVolume{37}
\acmNumber{4}
\acmArticle{111}
\acmMonth{8}

\begin{document}

\title{A Framework for Controllable Multi-objective Learning with Annealed Stein Variational Hypernetworks}

\author{Minh-Duc Nguyen}
\email{duc.nm2@vinuni.edu.vn}
\affiliation{%
  \institution{College of Engineering and Computer Science, VinUniversity}
  \city{Hanoi}
  \state{Hanoi}
  \country{Vietnam}
}

\author{Dung D. Le}
\authornote{Corresponding Author.}
\email{dung.ld@vinuni.edu.vn}
\affiliation{%
  \institution{College of Engineering and Computer Science, VinUniversity}
  \city{Hanoi}
  \state{Hanoi}
  \country{Vietnam}
}








\acmArticleType{Research}
\acmCodeLink{https://github.com/borisveytsman/acmart}
\acmDataLink{htps://zenodo.org/link}
\acmContributions{BT and GKMT designed the study; LT, VB, and AP
  conducted the experiments, BR, HC, CP and JS analyzed the results,
  JPK developed analytical predictions, all authors participated in
  writing the manuscript.}
\begin{abstract}
Pareto Set Learning (PSL) is an efficient approach for approximating the complete set of optimal solutions in Multi-objective Learning (MOL). By learning a set of solutions that map to a dense Pareto front in objective space, PSL aims to recover the underlying Pareto set. However, most existing methods focus heavily on convergence toward optimality, often neglecting solution diversity.
To explicitly balance convergence and diversity, we propose Stein Variational Hypernetwork for MOL (SVH-MOL), a novel framework that incorporates Stein variational updates into Pareto set learning. SVH-MOL updates solutions via two complementary components: (i) a driving term that guides particles toward the Pareto set, and (ii) a repulsive term that encourages diversity among solutions. These two terms inherently compete, making stable and effective learning challenging. To address this issue, we introduce an annealing schedule that adaptively controls the relative strength of each term during training. Extensive experiments on synthetic multi-objective benchmarks and real-world multi-task learning problems demonstrate that SVH-MOL achieves a better trade-off between convergence and diversity than existing methods.
\end{abstract}

\maketitle

\section{Introduction}
Multi-objective optimization (MOO) is a critical area of research in the field of optimization, focusing on problems that involve multiple conflicting objectives. Unlike traditional single-objective optimization, where the goal is to find a optimal solution(s) for a single objective function, MOO seeks to find a set of solutions that represent the trade-offs between the different objectives.MOO has been adopted across many domains, from energy system design \cite{Marler2004} to healthcare treatment planning \cite{Craft2006}, demonstrating its versatility in balancing conflicting objectives. In machine learning, multi-objective problems play a significant role in various applications, including recommender systems \cite{milojkovic2019multi,jannach2022multi, zaizi2023multi}, where they help balance multiple conflicting criteria, and multi-task learning \cite{sener2018multi,crawshaw2020multi}, where they optimize performance across multiple related tasks simultaneously. Furthermore, addressing multi-objective problems in real-world scenarios can be computationally demanding \cite{lin2022pareto}, particularly in high-dimensional settings \cite{nguyen2024high}, as it involves assessing multiple conflicting objectives, which increases computational costs. Recently, training large language models (LLMs) has integrated MOO to efficiently transfer downstream tasks, which are related to multi-task fine-tuning \cite{liu2024mftcoder} and multi-objective alignment \cite{wu2023fine, zhou2024beyond}.

A common method for addressing multi-objective problems is to transform them into a single-objective problem using scalarization. This method assigns weights to each objective, representing their trade-offs and relative importance. The optimization process then finds a single solution corresponding to a specific weight vector. As a result, the disadvantage is how to choose the desired weight. MOEA/D \cite{4358754} (Multi-Objective Evolutionary Algorithm based on Decomposition) is a widely used multi-objective optimization algorithm that decomposes a multi-objective problem into several single-objective subproblems and optimizes them simultaneously. MOEA/D enhances solution diversity by decomposing the problem into multiple subproblems, allowing for a more distributed exploration of the solution space. 

A current survey \cite{chen2025gradient} has classified MOO algorithms for multi-task learning into three approaches following the number of optimal solutions:
\begin{itemize}
    \item Finding a single Pareto solution: A single optimal solution is obtained, representing a solution with the contribution of each objective appropriately, ensuring that the final solution is not overly biased toward any particular objective. This balance is typically achieved through techniques such as loss balancing \cite{liu2019end,kendall2018multi,lin2024smooth} and gradient-based adjustments \cite{sener2018multi,chen2018gradnorm,yu2020gradient}, which dynamically regulate the influence of each objective during the optimization process.
    \item Finding a finite set of Pareto solutions: This approach provides users with a diverse set of options to choose from based on their specific requirements. Generally, preference vector-based methods are used to divide the problem into subproblems, where each solution is found by optimizing a specific subproblem defined by a preference vector. Furthermore, methods such as Pareto Multi-Task Learning (PMTL) \cite{lin2019pareto} leverage this strategy by evenly distributing preference vectors across the objective space to capture different trade-off regions. Alternatively, algorithms like Exact Pareto Optimal (EPO) \cite{mahapatra2020multi} Search use techniques such as gradient descent with controlled ascent to precisely navigate the Pareto front and locate solutions that exactly satisfy the user-specified preferences. 
    \item Finding an infinite set of Pareto solutions: This approach seeks to compute the complete Pareto front—that is, a continuum of Pareto optimal solutions that are continuously and densely distributed in the objective space. The hypernetwork-based method \cite{navon2020learning, Hoang2023} obtains a good Pareto front by mapping the preference vector space directly to the optimal solution space. Moreover, this hypernetwork-based framework significantly reduces the need for training multiple separate models for various preference settings. Instead, a single model can be employed to generate a diverse set of solutions, making it especially suitable for complex multi-task learning scenarios where balancing objectives is critical.
\end{itemize}

In the pioneer work, MOO-SVGD \cite{liu2021profiling}, Stein Variational Gradient Update (SVGD) is integrated to find a finite Pareto set. However, this adaptation suffers from several key weaknesses. The gradient descent direction is not in control, as it follows the Multiple Gradient Descent Algorithm (MGDA). It might be harder for MGDA to reach the global optimal region than other methods. Many weak Pareto points are obtained, which are not the desirable trade-off in MOO. Furthermore, MOO-SVGD in multi-task learning is not feasible. In fact, a large number of samples in training hinders the parameter optimization efficiency because of generating many architectures for all samples together. In other study, PHN-HVI \cite{Hoang2023}, the authors leverage hypervolume maximization \cite{deist2021multiobjectivelearningpredictpareto}  to derive the optimal solution while incorporating a penalty function to promote diversity among the resulting Pareto solutions. Using Hypervolume maximization as a gradient descent direction helps to find the optimal solution, but similar to MGDA, it tends to move towards a single optimal point. Therefore, in PHN-HVI, the incorporation of a penalty function promotes a more diverse spread of solutions along the Pareto front. However, this diversification can come at the cost of convergence precision, potentially steering the optimization process away from the most optimal points in the objective space.

To tackle the above issues, in our research, we investigate the third approach discussed in the context of Pareto set learning methods, which involves utilizing hypernetwork-based strategies. We leverage a hypernetwork to generate samples efficiently. We can adapt many strategies for controlling the Pareto front in training hypernet; thus, our approach is able to find a global Pareto set, which guarantees both convergence and diversity. For the hypernetwork-based approach, the issue about parameter optimization is mitigated, as the hypernetwork outputs the architecture for each sample, respectively.

We propose a novel approach for efficiently learning the complete Pareto set while capturing its inherent diversity, which employs Stein Variational Gradient Descent (SVGD) \cite{liu2016stein} to approximate the Pareto set. Our approach is designed to learn a continuous representation of the Pareto front, ensuring that all possible trade-offs are effectively captured. While the integration of Stein Variational Gradient Descent with hypernetworks was initalliy explored in our prior work SVH-PSL \cite{Nguyen_Dinh_Nguyen_Hoang_Le_2025} for expensive multi-objective optimization, this paper introduces several fundamental advancements that specifically address challenges in multi-objective learning: 
\begin{itemize}
    \item\textbf{Novel Annealing Framework for MOL}: We propose a specialized annealing schedule that dynamically balances convergence and diversity throughout training, unlike the static optimization in SVH-PSL designed for expensive black-box functions.
    \item \textbf{Controllable Scalarization Strategies}: We systematically investigate three distinct scalarization methods (Linear, Tchebyshev, Smooth Tchebyshev) within the SVGD framework and analyze their intricate interactions with the repulsive force term—an aspect unexplored in SVH-PSL.
    \item \textbf{Large-Scale Multi-Task Learning Focus}: While SVH-PSL targeted low-dimensional expensive optimization, SVH-MOL specifically addresses the scalability challenges in high-dimensional multi-task learning with deep neural networks, where parameter efficiency and training stability are paramount.
    \item \textbf{Comprehensive Theoretical and Empirical Analysis}: We provide extensive ablation studies on hyperparameter sensitivity, computational complexity, and convergence behavior, along with discussions on limitations and future directions—significantly expanding beyond the scope of SVH-PSL.
\end{itemize}

\section{Related Work}
\subsection*{Multi-objective Optimization (MOO)}
MOO is a fundamental challenge in machine learning, where it aims to find a set of optimal solutions that each solution represents the best trade-off between multiple conflicting objectives. Evolutionary algorithms are popular to find a set of diverse Pareto optimal solutions in a single run, such as VEGA \cite{schaffer2014multiple}, NSGA-III \cite{deb2013evolutionary}. This approach often faces
challenges in high-dimensional spaces and require significant computational resources. In contrast, gradient-based methods have emerged as more eﬀicient alternatives, particularly in differentiable settings. This method leverages the gradients of the objective functions to guide the search toward Pareto-optimal solutions. Scalarization techniques, such as the weighted sum method \cite{YANG2014197} and the Tchebyshev method \cite{steuer1983interactive}, provide an efficient approach for transforming multiple objectives into a single aggregated objective, enabling the use of gradient-based optimization methods.
\subsection*{Pareto Set Learning (PSL).} PSL emerges as a proficient means of approximating the entire Pareto front; it employs a hypernetwork \cite{ha2016hypernetworks} to learn a mapping between a preference vector and an optimal solution in a single training. Many studies proposed different gradient descent methods to optimize the combined-objective function, such as PHN-LS and PHN-EPO \cite{navon2020learning}. PHN-HVI \cite{Hoang2023} use maximizing the hypervolume indicator to guide the search for the optimal region. One framework for controllable Pareto front learning employs the completed scalarization functions proposed by \cite{tuan2024framework}. To exploit the geometric characteristics of the Pareto set in PSL, PSL-HV \cite{zhang2023hypervolume} is introduced, which derives Pareto solutions that closely align with the polar angle under mild conditions. In a recent study, PO-PSL \cite{haishan2024preference} focuses on the optimization of the preference vector instead of employing randomly generated vectors. This approach initiates a bilevel optimization problem in which it operates a preference-optimized term first. PO-PSL can lead to faster convergence to the true Pareto front. PSL has been effectively applied to optimization of multiple black-box functions \cite{lin2022pareto}, where it can efficiently suggest the next experiment to perform in a laboratory setting. The first work for Multi-Objective Reinforcement Learning is called PSL-MORL \cite{liu2025pareto}, which apply Pareto set learning to output a personalized policy network for each preference. 
\subsection*{Stein Variational Gradient Descent (SVGD).} SVGD \cite{liu2016stein} is a gradient method for approximating a target distribution, which was originally proposed for Bayesian inference. SVGD iteratively transports a set of particles to approximate a target distribution by minimizing the Kullback-Leibler (KL) divergence. However, as observed in \cite{zhuo2018message}, the particles of SVGD tend to collapse to modes of the target distribution, and this particle degeneracy phenomenon becomes more severe with higher dimensions.  This issue was explained in a previous study \cite{ba2021understanding}, the authors compared the SVGD update with the gradient descent on the maximum mean discrepancy (MMD). To address this challenge, annealed-SVGD \cite{d2021annealed} employs an annealing schedule to solve the inability of the particles to escape from local modes and the inefficiency in reproducing the density of the different regions. Selecting an appropriate kernel remains a key challenge, as the performance of the algorithm is highly dependent on the choice of kernel \cite{duncan2023geometry}. Multiple-kernel learning SVGD (MKL-SVGD)  \cite{ai2023stein} enables the combination of multiple kernels into a single, unified kernel, providing greater flexibility in the learning process. In MOO, the first study MOO-SVGD \cite{liu2021profiling} adapted SVGD to profile the Pareto front while maintaining diversity,
providing an eﬀicient alternative to traditional methods.
\section{Preliminaries}
\subsection{Multi-objective optimization}
Multi-objective optimization (MOO) refers to the process of simultaneously optimizing two or more conflicting objectives subject to a set of constraints. Unlike single-objective optimization, where a unique optimal solution exists, MOO typically results in a set of optimal trade-offs known as the Pareto front. 

Formally, a multi-objective optimization problem can be defined as:

\begin{equation}
\min_{x \in \mathcal{X}} \mathcal{F}(x) = \begin{bmatrix} f_1(x) \\ f_2(x) \\ \vdots \\ f_m(x) \end{bmatrix},
\end{equation}

where $\mathcal{F}(x) = [f_1(x), f_2(x), ..., f_m(x)]^T$ represents the $m$ objective functions, $\mathcal{F}(\cdot): \mathcal{X} \rightarrow \mathcal{Y}\subset \mathbb{R}^m$ and $\mathcal{X} \subset \mathbb{R}^n$ denotes the feasible decision space.
\begin{definition}[Dominance]
A solution $x^a$ dominates $x^b$ if:
    $f_j(x^a) \leq f_j(x^b),\forall j$ and $f_j(x^a) \neq f_j(x^b)$ for at least one $j$.
Denote $\mathcal{F}(x^a) \prec \mathcal{F}(x^b)$.
\end{definition}
\begin{definition}[Pareto Optimal Solution]
A solution $x^a$ is considered Pareto optimal if no other solution $x^b$ exists such that: $\mathcal{F}(x^b) \preceq \mathcal{F}(x^a).$
\end{definition}

\begin{definition}[Weakly Pareto Optimal Solution]
A solution $x^a$ is called weakly Pareto optimal if there is no solution $x^b$ such that: $ \mathcal{F}(x^b) \prec \mathcal{F}(x^a).$
\end{definition}
\begin{definition}[Pareto Set/Pareto Front]
The set of all Pareto optimal solutions is referred to as the Pareto set, denoted as $X_E$, and the corresponding objective space representation is known as the Pareto front, given by: $P_F = \mathcal{F}(X_E).$
\end{definition}
\subsection{Pareto Set Learning}
\begin{figure}[!ht]
    \centering
    \includegraphics[width=1.03\linewidth]{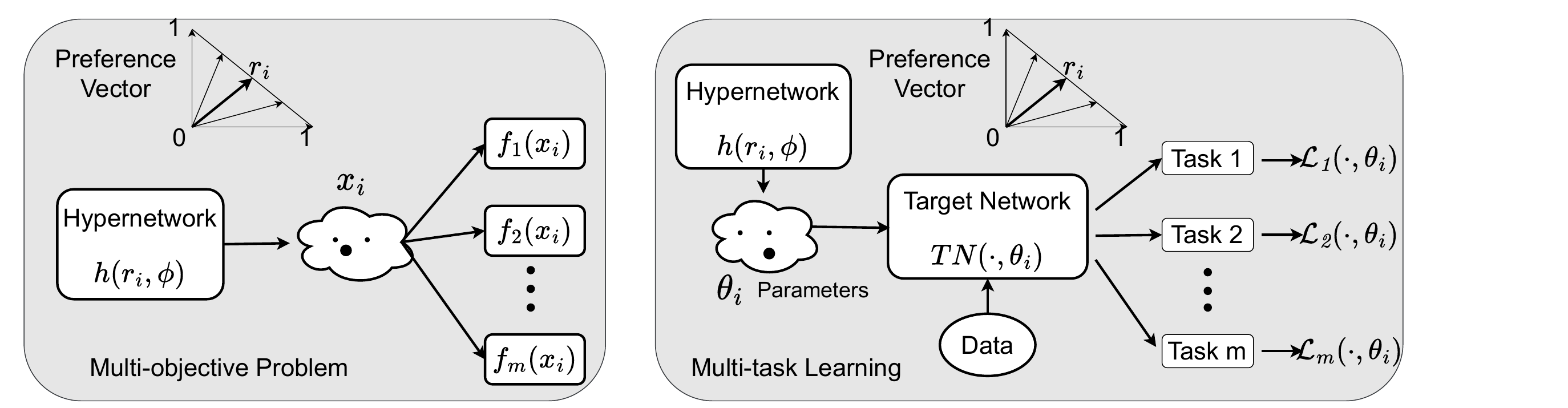}
    
\caption{The Pareto Set Learning (PSL) framework. \textbf{(Left)} In a general multi-objective problem, the hypernetwork $h(r_i, \phi)$ maps a preference vector $r_i$ to a solution $x_i$. \textbf{(Right)} In multi-task learning, the hypernetwork generates the parameters $\theta_i$ for a target network $TN(\cdot, \theta_i)$ that processes data for multiple tasks.}
    \label{fig: psl}
\end{figure}
Pareto Set Learning (PSL) offers an efficient approach to obtaining the entire Pareto set in a single training phase, contrasting with traditional methods that require varying hyperparameters to approximate only a few Pareto-optimal points. PSL employs a hypernetwork \cite{ha2016hypernetworks} to generate a diverse set of candidate solutions. During training, the hypernetwork's parameters are optimized by directly minimizing the multi-objective optimization (MOO) objective. Consequently, the trained hypernetwork produces a dense set of solutions that closely approximate the true Pareto front.

In our study, PSL is applied to multi-objective optimization, where the decision variable $x$ is generated by the hypernetwork. Additionally, in the context of multi-task learning, the hypernetwork is designed to output the full set of parameters for the target network responsible for solving the main tasks (Figure \ref{fig: psl}).

In multi-objective learning, our approach is based on a set of predefined preference vectors as a trade-off between the conflicting objectives, which is the input for the hypernetwork. We fomualate the optimize process as:
\begin{equation}
\begin{split}
    \phi^* = \arg\min_{\phi} \mathbb{E}_{r \sim \text{Dir}(\alpha)} \mathbf{s} \left( \mathcal{F} \left( \mathbf{x}_r \right), r \right)\\
     \text{s.t. } \mathbf{x}_r = \mathbf{h}(r, \phi^*) \in X_E, \, \mathbf{h} \left( \mathcal{P}, \phi^* \right) \equiv X_E
\end{split}
\end{equation}

where Hypernetwork \( h : \mathcal{P} \times \mathbb{R}^q \to \mathbb{R}^n \), \( q \) is the number of parameters of hypernetwork, and Dir\((\alpha)\) denoting the Dirichlet distribution and $r$ being the preference vector.
During the optimization process, we update the hypernetwork parameter $\phi$ as follows: 
\begin{equation}\label{updatephi}
\phi_{t+1} = \phi_t - \xi \nabla_\phi \mathbf{s} \left( \mathcal{F}  \left( \mathbf{x}_r \right), r \right) ,
\end{equation} 
where $\xi$ is the step size, $s(\cdot)$  is the complete function. Upon completing the optimization process, we obtain $\phi^*$, for each ray $r$, we derive the corresponding optimal solution $x_r = h(r,\phi^*)$.

In addition, for multi-task learning, our hypernetwork outputs the parameters $\theta$ of the target network $TN(\theta)$. In here, we obtain the loss function $\mathcal{L(\cdot,\theta})$ by fitting the data sample and the parameter $\theta$ to the target network. This problem follows:
\begin{equation}
    \phi^* = \arg \min_{\phi} \mathbb{E}_{r \sim Dir(\alpha)} s(\mathcal{L}(\cdot,\theta_ r), r)
\end{equation}
Here $\theta_r = h(r, \phi) $, the optimized process is similar to Formula (\ref{updatephi}).

Additionally, a complete function $s(\cdot)$ is employed to combine multiple objectives into a single objective for optimization. In previous studies, scalarization methods such as the linear or Chebyshev function have been used to identify optimal solutions corresponding to a given preference vector. Moreover, other studies such as PHN-HVI utilize the hypervolume indicator to construct an objective function that guides the optimization process. This function is designed to maximize the hypervolume of the Pareto front, thereby directing the updates toward more optimal regions in the objective space.
\subsection{Stein Variational Gradient Descent (SVGD)}
SVGD, a particle-based inference algorithm, is proposed to approximate the target distribution by applying the gradient information in the updating process. SVGD starts at a set of initial points $\{x_i^0\}_{i=1}^n \in \mathbb{R}^d$ drawn by a prior distribution $p(x)$, which is pushed to the target points $\{x^*_i\}_{i=1}^n$ belonging to the target distribution $q(x)$, following the iterative:
\begin{equation}
\begin{split}
    x_i^{t+1} \leftarrow x_i^t +\epsilon \phi^*(x_i^t), \forall i= 1,...,n, \\
     \phi^*_k = \arg\max_{\phi\ \in \mathcal{B}_k} \left\{ - \frac{d}{d\epsilon}\mathbf{K}\mathbf{L}(q_{[\epsilon \phi]} || p )\bigg|_{\epsilon=0}\right\}
     \end{split}
\end{equation} 
Where, $\phi^*$ is an optimal transform function chosen to maximize the decreasing rate of the KL divergence between the distribution of particles and the target $p$, and $q_{[\epsilon \phi]}$ is defined the distribution of the updated particles, and $\mathcal{B}_k$ is a unit ball of a reproducing kernel Hilbert space (RKHS) $\mathcal{H}^d_k := \mathcal{H}_k \times ... \times \mathcal{H}_k$, $\mathcal{H}_k$ is a Hilbert space associated with a positive definite kernel $k(x,x')$.
The authors have shown that $\phi^*$ can be expressed as a linear operator:
\begin{equation}
      \phi^*_k \propto \mathbb{E}_{x \sim q}[\mathcal{P}k(x, \cdot )] = \mathbb{E}_{x \sim q} \left[\nabla_{x} \log p(x)k(x,\cdot)
    + \nabla_{x}k(x,\cdot) \right]
\end{equation}

\section{Pareto Set Learning via Stein Variational Gradient Descent}
\subsection{Stein Variational Hypernetwork}

In this study, we develop a method for Pareto set learning that employs Stein Variational Gradient Descent to approximate the entire Pareto solution in multi-objective learning problems. Our method, SVH-MOL, bridges the theoretical framework of SVGD with practical multi-objective optimization through hypernetworks.

\subsubsection*{Theoretical Foundation and Connection to SVGD}

In traditional SVGD, particles are updated to approximate a target distribution $p(x)$ by minimizing the KL divergence. In our multi-objective context, we reinterpret this framework by considering the Pareto front as the target distribution in objective space. While the Pareto front is not a probability distribution in the conventional sense, we can define a target energy function that captures the desired properties of Pareto optimal solutions.

For a given preference vector $r$, we define the target energy function as:
\begin{equation}
E(\mathcal{F}) = \mathbf{g}(\mathcal{F}, r)
\end{equation}
where $\mathbf{g}$ is the scalarization function. The corresponding unnormalized target density becomes:
\begin{equation}
p(\mathcal{F}) \propto \exp(-E(\mathcal{F})) = \exp(-\mathbf{g}(\mathcal{F}, r))
\end{equation}

This formulation allows us to apply SVGD principles to push particles toward regions of low energy (good Pareto solutions) while maintaining diversity through repulsive forces.

\subsubsection*{Hypernetwork-based Particle Generation}

We employ a hypernetwork to generate initial particles, providing flexibility in adjusting particle count. Within the SVGD framework, each particle's movement is influenced by repulsive forces from neighboring particles. As particle count increases, these repulsive interactions intensify, leading to faster convergence and more diverse solution distribution across the Pareto front.
Building upon \cite{Nguyen_Dinh_Nguyen_Hoang_Le_2025}, SVH-MOL samples initial particles using a hypernetwork with predefined preference vectors as trade-off weights between objectives. We sample $n$ preference vectors $r_i$ and feed them to the hypernetwork $\mathbf{h}(\cdot,\phi)$ to generate solutions $\{\mathbf{x}_{r_i}\}_{i=1}^n$. For problems with $m$ objectives, the objective functions are represented as $\mathcal{F}_i(\mathbf{x}_{r_i}) = [f_1(\mathbf{x}_{r_i}), ..., f_m(\mathbf{x}_{r_i})]$, where each $\mathcal{F}_i$ is considered a particle.

The SVGD update rule modifies the hypernetwork parameter update as follows:
\begin{equation}
\phi_{t+1} = \phi_t - \xi \sum_{i=1}^n \sum_{j=1}^n \nabla_\phi\mathbf{g} \left( \mathcal{F}_i, r_i\right) \mathbf{k}(\mathcal{F}_i, \mathcal{F}_j) - \alpha\nabla_\phi\mathbf{k}(\mathcal{F}_i, \mathcal{F}_j)
\label{updatephi_svgd}
\end{equation}

where $\nabla_\phi\mathbf{g}(\mathcal{F}_i, r_i ) = \frac{\partial \mathbf{g}(\mathcal{F}_i, r_i)}{\partial \mathcal{F}_i} \cdot \frac{\partial \mathcal{F}_i}{\partial \phi}$ represents the chain rule through the hypernetwork, and $\mathbf{k}(\mathcal{F}_i, \mathcal{F}_j)$ is the Gaussian kernel matrix.

\begin{itemize}
    \item The term $\nabla_\phi\mathbf{g}(\mathcal{F}_i) \mathbf{k}(\mathcal{F}_i, \mathcal{F}_j)$ acts as the \emph{driving force}, guiding solutions toward optimal regions by following the gradient of the energy function.
    \item The term $\nabla_\phi\mathbf{k}(\mathcal{F}_i, \mathcal{F}_j)$ serves as the \emph{repulsive force}, maintaining diversity by preventing particle collapse.
\end{itemize}

This formulation ensures that solutions simultaneously converge to the Pareto front while maintaining good distribution across it.

\subsection{Controllable Pareto Set Learning}
In traditional Stein Variational Gradient Descent (SVGD), a discrepancy measure known as the Stein discrepancy is used to quantify the difference between the current and target distributions. In a more recent study, \cite{ba2021understanding} introduced Mean Discrepancy Descent (MMD descent), a novel approach that replaces the original driving force in SVGD with one based on the Maximum Mean Discrepancy. Their work highlights a key limitation of SVGD, variance collapse, when approximating high-dimensional distributions. To address this issue, they modify the gradient update formulation, suggesting that for different problems, it may be beneficial to design a problem-specific driving force to guide particles toward better optimal solutions.

In the context of multi-objective learning, the Stein Variational Hypernetwork (SVH) is proposed to approximate the entire Pareto set. This approach leverages Stein's updating rule while utilizing a hypernetwork to generate candidate solutions. The idea of using a learnable hypernetwork to produce the Pareto set was introduced by \cite{tuan2024framework}, who proposed the Controllable Pareto Front Learning (CPF) framework. CPF trains a single hypernetwork whose parameters are optimized using principles from scalarization theory.

To enable controllable Pareto set learning in SVH, we incorporate three scalarization methods: Linear, TChebyshev, and Smooth TChebyshev. Each scalarization method modifies the driving force in SVH by computing the value $\nabla_\phi g(\mathcal{F})$ as the gradient of the respective scalarization function. Each method induces a distinct optimization behavior, determined by the direction of its gradient. By experimenting with different scalarization, we aim to understand the impact of the repulsive term on SVH and how it contributes to the diversity and completeness of the resulting Pareto Front.
\subsection*{Linear Scalarization Function (LS)}
In multi-objective optimization, Linear Scalarization (LS) is a straightforward approach that combines multiple objectives into a single objective by computing a weighted sum. The formula is defined as: 
\begin{equation}
    \mathbf{g}(\mathcal{F}_i, r_i)=\sum_{j=1}^m r_{i,j} f_j(x_{r_i})
\end{equation}
The weakness of this approach is that LS does not work well in non-convex problems. In these problems, the set of solutions lies on the convex envelope of the Pareto set. Therefore, the solution found is not aligned between the preference vector space and the objective space. In SVH, this disadvantage raises an interesting issue: whether the repulsive term can push the optimal solution into the non-convex region.
\subsection*{Tchebyshev Function (TCH)}
The Chebyshev function is popular for many problems as it can produce the complete Pareto set in both non-convex and convex problems. It scalarizes multiple objectives by considering the maximum weighted deviation from the ideal point, which helps guide the optimization process even in regions where linear scalarization fails. The function is typically defined as:

\begin{equation}
    \mathbf{g}(\mathcal{F}_i,r_i) = \max_{j=1,\ldots,m} r_{i,j} |f_j(x_{r_i}) - z_j^*|
\end{equation}

where $z^*$ is the ideal point. This formulation encourages solutions that minimize the worst-case deviation from the ideal point, making it well-suited for exploring non-convex regions of the Pareto front. However, the max operator introduces non-smoothness, which can hinder gradient-based optimization methods.

\subsection*{Smooth Tchebyshev Function (STCH)}
To overcome the non-smooth nature of the standard Tchebyshev function, the Smooth Tchebyshev function \cite{lin2024smooth} replaces the max operator with a smooth approximation, such as the softmax or log-sum-exp function. A commonly used form is:

\begin{equation}
    \mathbf{g}(\mathcal{F}_i, r_i) = \frac{1}{\mu} \log \left( \sum_{j=1}^m \exp\left( \mu r_{i,j} |f_j(x_{r_i}) - z_j^*| \right) \right)
\end{equation}

where $\mu > 0$ controls the smoothness. As $\mu \to \infty$, the function approaches the standard Tchebyshev function. This smooth approximation enables the use of gradient-based optimization while retaining the desirable properties of the Tcchebyshev scalarization, such as its ability to explore non-convex regions and generate diverse solutions across the Pareto front.

\subsection{Diversity-Promoting via Annealed SVH-MOL}
\label{sec: diversity}
\begin{figure}[!t]
    \centering
    \includegraphics[width=0.6\linewidth]{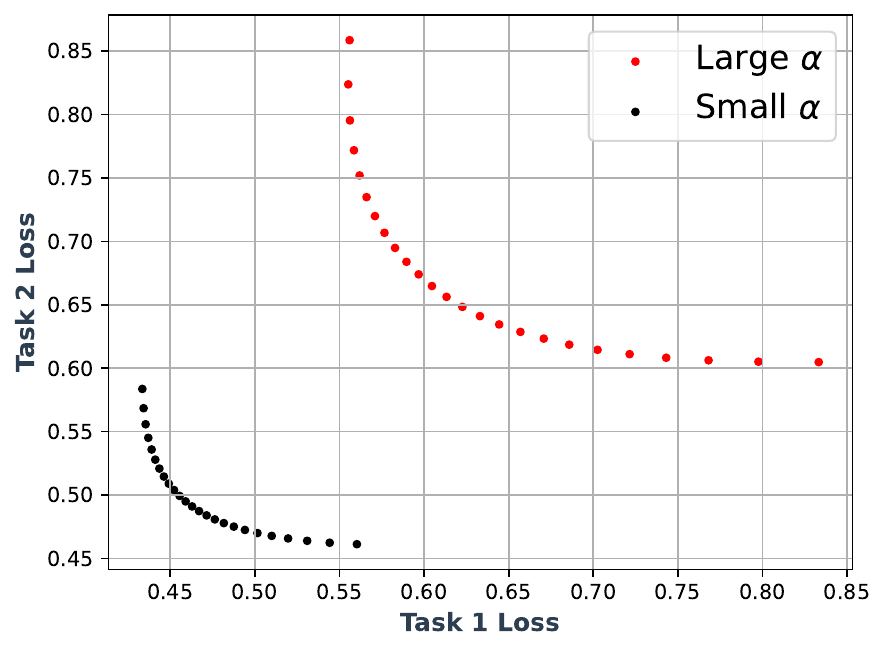}
\caption{Illustration of the trade-off controlled by the diversity hyperparameter $\alpha$ (see Eq. 11). A \textbf{large $\alpha$} (red dots) prioritizes the repulsive force, leading to a diverse but suboptimal Pareto set (poor convergence). A \textbf{small $\alpha$} (black dots) results in a well-converged front but with potentially less spread.}
    \label{fig: alpha}
\end{figure}

In our optimization process, the gradient updates are influenced by two conflicting components. The first term acts as a driving force, guiding the solutions toward the optimal region. In contrast, the second term serves as a repulsive force, promoting diversity by preventing the solutions from collapsing onto one another along the Pareto front. This issue highlights the trade-off between convergence and diversity. While a Pareto front may exhibit good convergence, it does not necessarily guarantee diversity. In our experiments, we observed that prioritizing the repulsive term often leads to a Pareto set with poor optimality (Figure \ref{fig: alpha}).
In previous work, \cite{liu2021profiling} discussed the hyperparameter $\alpha$ in Formula (\ref{updatephi_svgd}), highlighting its influence on the diversity of the Pareto front. However, they were unable to effectively control this effect. In a recent study, \cite{d2021annealed} also investigated this challenge and reported a variance collapse issue. They observed that particles in SVGD tend to collapse into a few local modes, a phenomenon strongly influenced by the initial particle distribution. To address this, they proposed an annealing schedule that gradually increases the weight of the driving force term. This strategy encourages the initial particles to spread across diverse regions at the beginning, guiding them toward a more diverse distribution throughout the optimization process.

\begin{equation}
\phi_{t+1} = \phi_t - \xi \sum_{i=1}^n \sum_{j=1}^n \gamma (t) \nabla_\phi \mathbf{g} \left( \mathcal{F}_i \right) \mathbf{k}(\mathcal{F}_i, \mathcal{F}_j) - \alpha\nabla_\phi\mathbf{k}(\mathcal{F}_i, \mathcal{F}_j)
\label{updatephi_svgd_annealing}
\end{equation} 
Where $\gamma(t)$ is a coefficient that controls the influence of the driving term on the overall gradient update. \cite{d2021annealed} varies $\gamma(t)$ in the interval $[0,1]$ with an appropriate schedule; the process consists of two phases. The first is an exploratory phase, where a predominant repulsive force drives the points away from their initial positions, promoting broad coverage of the target distribution’s support. The second is an exploitative phase, in which the driving force becomes dominant, guiding the points to concentrate around the various modes of the distribution.

Motivated by this approach, we propose a novel method for controlling the influence of the driving term. Rather than applying the scheduling function $\gamma(t)$ throughout the entire training process, we restrict its use to only the initial few epochs. Our rationale is that once the initial particles are sufficiently dispersed across different regions of the distribution, continued application of the pushing schedule in later epochs becomes unnecessary. Thus, our $\gamma(t)$ is formulated as:
\begin{equation}
    \gamma(t) = \frac{mod(t, T_j)}{T_j}, T_{j+1} = \tau T_j  \label{eq: gama}
\end{equation}
Here, we construct $\gamma(t)$ to vary over an initial long period of length $T_0$, after which the length of each subsequent period $T_i$ decreases according to a decay factor $\tau$.

The motivation for our proposed approach stems from the observation that, within our framework, the SVGD gradient updates are highly sensitive to the parameters of the hypernetwork. Consequently, applying a strong repulsive force in the later training epochs can lead to instability in the convergence process, particularly near the optimal region.

\begin{algorithm}[!t]
\caption{Annealed Stein Variational Hypernetwork for Multi-Objective Learning (A-SVH-MOL)}
\label{alg:svh-mol}
\begin{algorithmic}[1]
\Require Objective functions $\mathcal{F} = \{f_1, \ldots, f_m\}$; number of particles $n$; learning rate $\xi$; diversity hyperparameter $\alpha$; annealing parameters $T_0$, $\tau$; number of training epochs $E$
\Ensure Trained hypernetwork parameters $\phi^*$
\State Initialize hypernetwork parameters $\phi$
\For{epoch $t = 1$ to $E$}
    \State Sample batch of $n$ preference vectors $\{r_i\}_{i=1}^n \sim \text{Dir}(\alpha)$
    \State Generate particles via hypernetwork: $\{x_i\}_{i=1}^n = h(\{r_i\}_{i=1}^n, \phi_t)$
    \State Compute objective values: $\{F_i\}_{i=1}^n = \{\mathcal{F}(x_i)\}_{i=1}^n$
    \State Calculate annealing coefficient $\gamma(t)$ using Eq. (\ref{eq: gama})
    \State Initialize gradient accumulator: $\nabla_{\phi} I \leftarrow 0$
    \For{each particle $i = 1$ to $n$}
        \State Compute $\nabla_{\phi} \mathbf{g}(\mathcal{F}_i)$ \Comment{Gradient of scalarization function w.r.t. $\phi$}
        \For{each particle $j = 1$ to $n$}
            \State Compute kernel: $k_{ij} = \exp\left(-\frac{\|F_i - F_j\|^2}{2h^2}\right)$
            \State Compute driving force: $\text{drive} = \gamma(t) \cdot \nabla_{\phi} \mathbf{g}(\mathcal{F}_i) \cdot k_{ij}$
            \State Compute repulsive force: $\text{repulse} = \alpha \cdot \nabla_{\phi} k_{ij}$
            \State Accumulate gradient: $\nabla_{\phi} I \leftarrow \nabla_{\phi} I + \text{drive} + \text{repulse}$
        \EndFor
    \EndFor
    \State Update parameters: $\phi_{t+1} \leftarrow \phi_t - \xi \cdot \nabla_{\phi} I$
\EndFor
\State \textbf{return} $\phi^* \leftarrow \phi_E$
\end{algorithmic}
\end{algorithm}

\section{Experiments}

\subsection{Evaluation Metrics}
\textbf{Mean Euclidean Distance (MED).}\cite{tuan2024framework} To evaluate the effectiveness of the controllable Pareto front, which means that the hypernetwork generates the optimal Pareto solution, aligning with the preference vector. We measure the error between the obtained Pareto solutions by the hypernetwork and the true Pareto solutions through calculating the mean of the Euclidean distance. The metric MED is defined as:
\begin{equation}
    MED(\mathcal{F}^*, \mathcal{\hat F}) = \frac{1}{n} (\sum_{i=1}^n||\mathcal{F}^*_i - \mathcal{\hat F}_i||_2)
\end{equation}

Here $\mathcal{F^*} = \{\mathcal{F^*}_1,..., \mathcal{F^*}_n \}$ are the true Pareto solutions and the generated Pareto solutions $\mathcal{\hat F} = \{\mathcal{\hat F}_1,...,\mathcal{\hat F}_n\}$ corresponding to preference vector set $r \in \mathbb{R}^m_{>0}: \sum_i^mr_i=1$. $n$ is the number of rays that we consider for validation. A lower MED value indicates that the approximate solution is closer to the true Pareto front.

\textbf{Hypervolume (HV).} Hypervolume \cite{zitzler1999multiobjective} is a metric to figure out both convergence and diversity in the obtained solution. For some problems for which we lack the ground truth Pareto set, HV is an effective way to validate the quality of obtained solutions. The higher HV value reflects that the method is better. Given a set of $n$ points $y = \{y^{(i)} | y^{(i)} \in \mathbb{R}^m; i=1,\dots, n\}$ and a reference point $\rho \in\mathbb{R}^m$, the Hypervolume of $y$ is measured by the region of non-dominated points bounded above by $y^{(i)} \in y$, then the hypervolume metric is defined as follows:
    \begin{equation}
        HV(y) = VOL\left(\underset{y^{(i)} \in y, y^{(i)} \prec \rho}{\bigcup}\displaystyle{\Pi_{i=1}^n}\left[y^{(i)},\rho_i\right]\right)
    \end{equation}
 where $\rho_i$ is the i$^{th}$ coordinate of the reference point $\rho$ and $\Pi_{i=1}^n\left[y^{(i)},\rho_i\right]$ is the operator creating the n-dimensional hypercube from the ranges $\left[y^{(i)},\rho_i\right]$.

\textbf{$\Delta $-Spread}  \cite{deb2002fast} metric is a widely used measure for evaluating the diversity and distribution quality of approximated Pareto fronts in multi-objective optimization. This quantifies both the uniformity of solution distribution and the coverage of extreme points along the Pareto front.

The $\Delta $-Spread metric is defined as:

\begin{equation}
\Delta = \frac{d_f + d_l + \sum_{i=1}^{n-1} |d_i - \bar{d}|}{d_f + d_l + (n-1)\bar{d}}
\end{equation}

\noindent where:

\begin{itemize}
    \item $n$: Number of solutions in the approximated Pareto front
    \item $d_i$: Euclidean distance between consecutive solutions $i$ and $i+1$ after sorting
    \item $\bar{d}$: Average of consecutive distances, $\bar{d} = \frac{1}{n-1}\sum_{i=1}^{n-1} d_i$
    \item $d_f, d_l$: Distances to the extreme points of the true Pareto front
\end{itemize}
A low $\Delta $-Spread value reflects a good Pareto set with more uniformity in the objective space.


\subsection{Multi-objective Problems}
In this study, we conduct extensive experiments to demonstrate our framework's effectiveness in many use cases. We compare the performance of our SVH-MOL with the baseline PSL methods: PHN-EPO, PHN-LS \cite{navon2020learning}, PHN-TCH (TChebyshev), and PHN-STCH (Smooth Tchebyshev) \cite{lin2024smooth}.
\subsubsection{Synthetic Problems}
To validate the SVH-MOL's ability in towarding the optimal solution, we compare the MED on three synthetic tests: ZDT1, ZDT2 \cite{10.1162/106365600568202}.  These ZDT problems have two objectives and 30 variables with a convex (ZDT1) and non-convex (ZDT2) Pareto-optimal set. In the optimization process, we train the hypernetwork with 20000 iterations and the hyperparameter $\alpha = 1e-5$. We compute MED value with the number of preference vectors for testing from 30 to 600 rays $r_i$.
\begin{table}[!t]
\centering
\resizebox{0.5\textwidth}{!}{
\begin{tabular}{c|| c c c c }
\toprule
Rays & LS & Chebyshev & Smooth Chebyshev & SVH (ours)  \\
\midrule
\multicolumn{5}{c}{\textbf{ZDT1}} \\
\midrule
30  & 2.8184e-3 & 8.9748e-3 & 7.2197e-3 & \textbf{2.5042e-3}  \\
50  & 2.0595e-3 & 8.4050e-3 & 6.9985e-3 &\textbf{2.0582e-3} \\
100 & 2.0372e-3 & 7.7228e-3 & 7.0542e-3 & \textbf{2.0300e-3} \\
300 & 2.0403e-3 & 7.2580e-3 & 7.0194e-3 & \textbf{2.0212e-3} \\
600 & 1.9957e-3 & 7.2239e-3 & 6.9931e-3 & \textbf{1.8053e-3}   \\
\midrule
\multicolumn{5}{c}{\textbf{ZDT2}} \\
\midrule
30  & 7.0711e-1 & 6.0691e-3 & 2.4362e-3        & \textbf{2.0653e-3}        \\
50  & 7.0711e-1 & 6.1505e-3 & 2.0391e-3        & \textbf{2.0001e-3}        \\
100 & 7.0711e-1 & 5.9876e-3 & 2.0329e-3         & \textbf{2.0010e-3}        \\
300 & 7.0711e-1 & 6.0401e-3 & 2.0487e-3      & \textbf{1.9995e-3}        \\
600 & 7.0711e-1 & 5.9831e-3 & 2.0199e-3      & \textbf{1.9832e-3}        \\

\bottomrule
\end{tabular}
}
\caption{Mean Euclidean Distance (MED) comparison on ZDT1 (convex) and ZDT2 (non-convex) synthetic problems. Our vanilla SVH-MOL (without annealing) is compared against baseline scalarization methods (LS, TCH, STCH) across varying numbers of test rays. Lower values indicate better alignment with the true Pareto front.}
\label{tab:performance-MED}
\end{table}


In table \ref{tab:performance-MED}, our SVH-MOL framework demonstrates that the performance of MED on synthetic tests outperform baseline methods. The main point is that SVH-MOL ensures convergence by following scalarization methods, while the repulsive term pushes the solutions toward better alignment between the preference vectors and the Pareto-optimal points. In this experiment, we set a small value $\alpha$ to avoid the solution being pushed away from the optimal region. Moreover, in this result, we report only the vanilla SVH-MOL without the annealing schedule. Since the proposed annealing version of SVH-MOL encourages the discovery of a more diverse Pareto set, the resulting solutions may be suboptimal on synthetic problems.

\subsubsection{Real-world Problems}
We evaluate all methods on 11 RE problems (real-world engineering design problems) \cite{tanabe2020easy}, which involve either two or three objectives. These problems present both convex and non-convex Pareto fronts, and some exhibit highly complex Pareto-optimal sets. Therefore, we aim to assess our framework on these benchmarks to evaluate its ability to approximate diverse and challenging Pareto fronts under realistic conditions.

\begin{table}[!ht]
\centering
\resizebox{0.7\textwidth}{!}{
\begin{tabular}{l ccc c}
\toprule
& \multicolumn{3}{c}{Baselines (PHN)} & \multicolumn{1}{c}{A-SVH-MOL (Ours)} \\
\cmidrule(r){2-4} \cmidrule(r){5-5}
Problem & LS & TCH & STCH & Best Result \\
\midrule
RE21 & 0.8325 $\pm$ 0.0000 & 0.8330 $\pm$ 0.0001 & \textbf{0.8344 $\pm$ 0.0001} & \underline{0.8336 $\pm$ 0.0004}  \\
RE22 & 0.3867 $\pm$ 0.1176 & 0.4100 $\pm$ 0.0517 & \underline{0.4152 $\pm$ 0.0602} & \textbf{0.5537 $\pm$ 0.0135} \\
RE23 & 0.0574 $\pm$ 0.0269 & 0.0615 $\pm$ 0.0323 & \underline{0.0629 $\pm$ 0.0340} & \textbf{0.8955 $\pm$ 0.0662} \\
RE24 & \textbf{1.1636 $\pm$ 0.0001} & 1.1628 $\pm$ 0.0002 & \underline{1.1632 $\pm$ 0.0001} & \textbf{1.1636 $\pm$ 0.0000} \\
RE31 & \textbf{1.3310 $\pm$ 0.0000} & 1.3309 $\pm$ 0.0000 & 1.3309 $\pm$ 0.0000 & \textbf{1.3310 $\pm$ 0.0000} \\
RE32 & 1.3295 $\pm$ 0.0000 & \underline{1.3297 $\pm$ 0.0000} & 1.3295 $\pm$ 0.0000 & \textbf{1.3300 $\pm$ 0.0003} \\
RE33 & \underline{0.9948 $\pm$ 0.0037} & 0.9907 $\pm$ 0.0027 & 0.9892 $\pm$ 0.0004 & \textbf{1.0158 $\pm$ 0.0007} \\
RE34 & 0.8430 $\pm$ 0.1248 & \textbf{0.9951 $\pm$ 0.0022} & \underline{0.9932 $\pm$ 0.0053} & 0.9103 $\pm$ 0.0543  \\
RE35 & 0.1865 $\pm$ 0.0627 & \underline{1.0914 $\pm$ 0.0317} & 1.0793 $\pm$ 0.0229 & \textbf{1.2934 $\pm$ 0.0008} \\
RE36 & \underline{1.0187 $\pm$ 0.0001} & 1.0186 $\pm$ 0.0000 & \textbf{1.0188 $\pm$ 0.0000} & 1.0184 $\pm$ 0.0003 \\
RE37 & 0.7870 $\pm$ 0.0545 & 0.8264 $\pm$ 0.0005 & \textbf{0.8395 $\pm$ 0.0005} & \underline{0.8380 $\pm$ 0.0008} \\
\bottomrule
\end{tabular}
}
\caption{Hypervolume (HV) comparison on 11 real-world (RE) engineering problems. Our A-SVH-MOL (using the best-performing scalarization) is compared against the baseline PHN methods (PHN-LS, PHN-TCH, PHN-STCH). Results are shown as mean $\pm$ standard deviation. Higher HV values indicate better convergence and diversity.}
\label{tab:RE_best_A-SVH}
\end{table}

\begin{table}[!ht]
\centering
\resizebox{0.7\textwidth}{!}{
\begin{tabular}{l ccc c}
\toprule
& \multicolumn{3}{c}{Baselines (PHN)} & \multicolumn{1}{c}{A-SVH-MOL (Ours)} \\
\cmidrule(r){2-4} \cmidrule(r){5-5}
Problem & LS & TCH & STCH & Best Result \\
\midrule
RE21 & 0.7734 $\pm$ 0.0067 & \underline{0.2956 $\pm$ 0.0108} & 0.2631 $\pm$ 0.0030 & \textbf{0.2406 $\pm$ 0.0160} \\
RE22 & 0.9887 $\pm$ 0.1806 & \underline{0.9650 $\pm$ 0.2776} & 0.9884 $\pm$ 0.2882 & \textbf{0.8881 $\pm$ 0.0549} \\
RE23 & 1.0156 $\pm$ 0.0149 & \underline{0.9838 $\pm$ 0.0359} & 1.0119 $\pm$ 0.0076 & \textbf{0.9190 $\pm$ 0.0534} \\
RE24 & 0.9903 $\pm$ 0.0127 & \underline{1.0116 $\pm$ 0.0051} & 1.0398 $\pm$ 0.0008 & \textbf{0.8105 $\pm$ 0.0582} \\
RE31 & \textbf{1.0617 $\pm$ 0.1737} & 1.4388 $\pm$ 0.1091 & \underline{1.2245 $\pm$ 0.1701} & 1.3147 $\pm$ 0.1968 \\
RE32 & \underline{1.0006 $\pm$ 0.0321} & 1.0126 $\pm$ 0.0030 & \textbf{0.9994 $\pm$ 0.0005} & 1.0089 $\pm$ 0.0039 \\
RE33 & 1.2229 $\pm$ 0.0207 & 1.0226 $\pm$ 0.0837 & \textbf{0.8553 $\pm$ 0.0573} & \underline{0.9465 $\pm$ 0.1077} \\
RE34 & {\ul 1.0259 $\pm$ 0.2493} & 1.3484 $\pm$ 0.0462 & 1.4253 $\pm$ 0.0173 & \textbf{1.0217 $\pm$ 0.2179} \\
RE35 & \textbf{1.0547 $\pm$ 0.0234} & 1.2921 $\pm$ 0.2300 & 1.4799 $\pm$ 0.2472 & \underline{1.0561 $\pm$ 0.0540} \\
RE36 & 0.8572 $\pm$ 0.0236 & \textbf{0.8258 $\pm$ 0.0412} & 0.9218 $\pm$ 0.0519 & \underline{0.8382 $\pm$ 0.0647} \\
RE37 & 1.4470 $\pm$ 0.0712 & 0.8663 $\pm$ 0.0151 & \underline{0.7936 $\pm$ 0.0137} & \textbf{0.7523 $\pm$ 0.0280} \\
\bottomrule
\end{tabular}
}
\caption{$\Delta$-Spread ($\Delta$) metric comparison on real-world (RE) problems. This metric evaluates the uniformity and distribution of solutions. Our A-SVH-MOL is compared against baseline PHN methods. Results are shown as mean $\pm$ standard deviation. Lower values indicate a more uniform and well-distributed Pareto set.}
\label{tab:RE_spread}
\end{table}
In Table \ref{tab:RE_best_A-SVH}, SVH-MOL with the annealing schedule demonstrates superior performance, as A-SVH-MOL consistently achieves higher HV values compared to the baseline methods. Across each combination of SVH and a scalarization method, our proposed SVH-MOL generally obtains better results than the corresponding baselines.

Table \ref{tab:RE_spread} presents the $\Delta$-Spread metric, which quantifies both the uniformity of solution distribution and coverage of extreme points along the Pareto front. Our A-SVH-MOL achieves the best (lowest) $\Delta$-Spread values in \textbf{7 out of 11 problems} (RE21, RE22, RE23, RE24, RE32, RE35, and RE37), demonstrating superior distribution quality across diverse problem geometries. The consistent $\Delta$-Spread improvements across most problems highlight A-SVH-MOL's ability to maintain uniform solution distributions while achieving competitive convergence. This balanced performance is particularly valuable in engineering applications where both solution quality and coverage of the trade-off space are critical for decision-making.

\begin{figure}[!ht]
    \centering
    \includegraphics[width=0.5\linewidth]
    {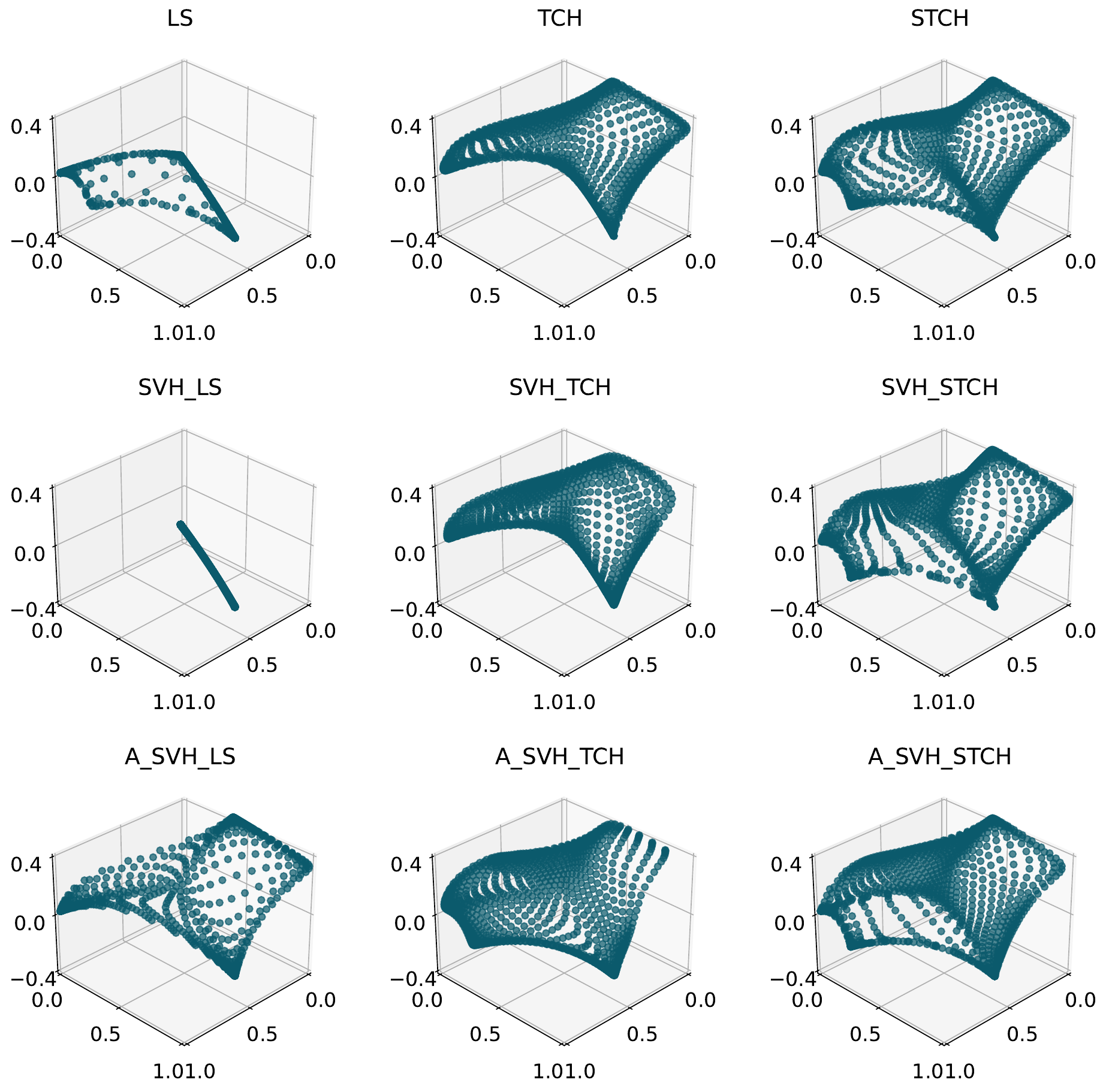}
    \caption{Visualization of Pareto fronts on the three-objective RE37 benchmark. The plots compare baseline PHN methods (top row) with our A-SVH-MOL methods (bottom row) using Linear (LS), Tchebyshev (TCH), and Smooth Tchebyshev (STCH) scalarization. Our A-SVH variants achieve superior coverage and diversity.}
    \label{fig: re37}
\end{figure}

Our SVH-MOL successfully captures the complete Pareto front, ensuring both convergence and diversity. In Figure \ref{fig: re37}, we visualize the Pareto front on RE37 with three objectives. The results highlight the effectiveness of our framework in handling complex Pareto sets. The Pareto fronts generated by A-SVH-\{LS, TCH, STCH\} demonstrate better coverage compared to the baselines. Notably, in A-SVH-LS and A-SVH-TCH, the obtained Pareto sets cover most of the optimal solution space, unlike PHN-LS or PHN-TCH, which only capture a limited portion. This confirms the robustness and adaptability of our approach. Overall, SVH-MOL leverages the repulsive force term to promote diversity and prevent premature convergence.


\subsection{Multi-task Learning}
In Multi-task learning, we compare our SVH-MOL performance with the baseline methods: PHN-LS, PHN-EPO, PHN-TCH, PHN-STCH, and PHN-HVI \cite{Hoang2023}. The multi-task experiments consist of two benchmarks: image classification (Multi-MNIST, Multi-Fashion, Multi-Fashion+ MNIST) \cite{lin2019pareto}, and Multi-Output Regression (SARCOS) \cite{vijayakumar2000sarcos}
\subsubsection*{Image Classification}
We train a hypernetwork to generate the complete set of parameters for a target network that performs image classification across three datasets: Multi-MNIST, Multi-Fashion, and Multi-MNIST+Fashion. Each dataset defines a two-objective multi-task learning problem, where the goal is to find a set of optimal Pareto solutions. Specifically, the tasks involve classifying two overlapping handwritten digits in Multi-MNIST, two overlapping fashion items in Multi-Fashion, and one handwritten digit and one fashion item in Multi-MNIST+Fashion. We use both the LSTM \cite{10.1162/neco.1997.9.8.1735} layer and the MLP layer in our core hypernetwork, and we employ a Multi-LeNet \cite{sener2018multi} for the target network. In the training process, we set the number of preference vectors $n = 4$ and train for 500 epochs.

\begin{figure}[!ht]
    \centering
    \includegraphics[width=1\linewidth]{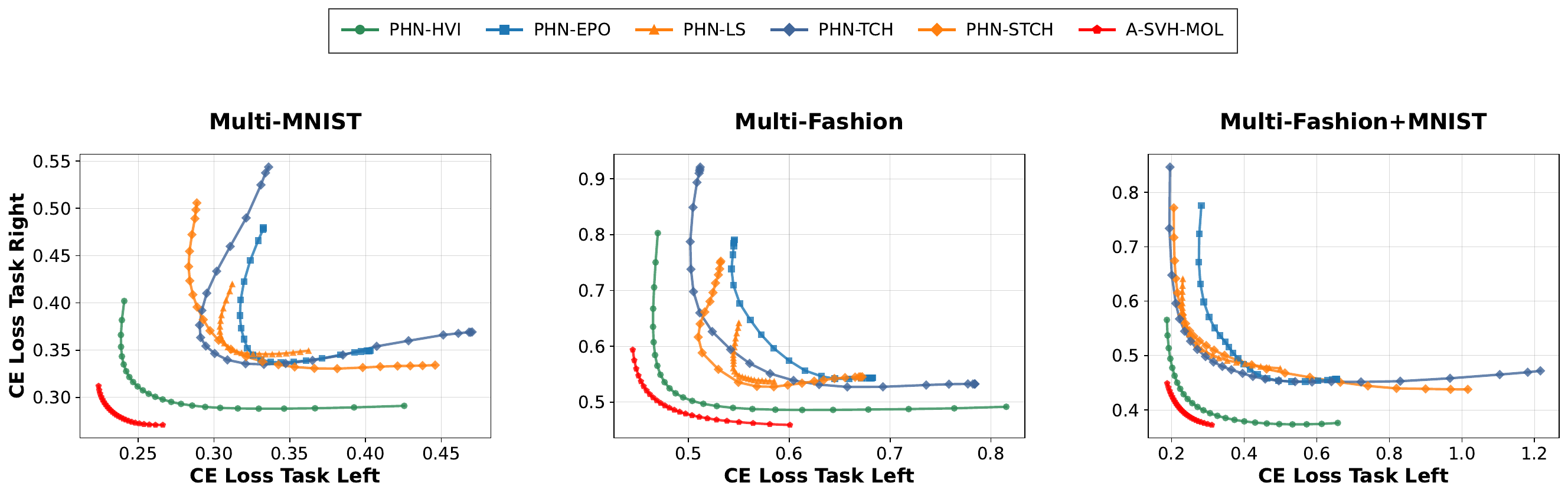}
    \caption{Generated Pareto fronts on multi-task image classification benchmarks: \textbf{(Left)} Multi-MNIST, \textbf{(Center)} Multi-Fashion, and \textbf{(Right)} Multi-MNIST+Fashion. Our A-SVH-MOL method (black line) is compared against several PHN baselines. In all three datasets, A-SVH-MOL achieves a superior Pareto front, representing a better trade-off between the two task losses.}
    \label{fig: image_cls}
\end{figure}

In table~\ref {tab: performance_comparison}, A-SVH-MOL presents outperformed results on both hypervolume and $\Delta-Spead$ value in comparison with baselines. This demonstrates that our framework preserves both the convergence and diversity in the Pareto set. In three datasets, A-SVH-MOL achieves high-mean and low-standard-difference results, indicating the stability of our framework. Specifically, these results show the harmony between the driving force and the repulsive force in our A-SVH-MOL. While the first term strongly pushes the solution set toward the optimal region, the second term maintains diversity and uniformity in the Pareto set. The Pareto fronts presented in figure~\ref{fig: image_cls} illustrates that A-SVH-MOL obtains a good curve. 
\begin{table}[!ht]
\centering
\resizebox{\columnwidth}{!}{%
\begin{tabular}{@{}l*{8}{c}@{}}
\toprule
\multirow{2}{*}{\textbf{Method}} & \multicolumn{2}{c}{\textbf{Multi-MNIST}} & \multicolumn{2}{c}{\textbf{Multi-Fashion}} & \multicolumn{2}{c}{\textbf{Fashion-MNIST}}  & \multicolumn{2}{c}{\textbf{SARCOS}} \\
\cmidrule(lr){2-3} \cmidrule(lr){4-5} \cmidrule(lr){6-7} \cmidrule(lr){8-9}
& \textbf{HV} & \textbf{$\Delta $-Spread}   & \textbf{HV} & \textbf{$\Delta $-Spread}   & \textbf{HV} & \textbf{$\Delta $-Spread}  & \textbf{HV} & \textbf{$\Delta $-Spread}    \\
\midrule
PHN-LS              & 2.8401 $\pm$ 0.0151 & 0.6400 $\pm$ 0.2609 & 2.1845 $\pm$ 0.0208 & 0.7064 $\pm$ 0.2592 &2.8069 $\pm$ 0.0236 & 0.4346 $\pm$ 0.0668& 0.7249 $\pm$ 0.0136  & 0.6628 $\pm$ 0.1964 \\
PHN-TCH       & 2.8391 $\pm$ 0.0187 &0.7711 $\pm$ 0.2213 & 2.1758 $\pm$ 0.0340 &0.8998 $\pm$ 0.1332 & 2.8041 $\pm$ 0.0206 &0.6466 $\pm$ 0.0764 &0.7093 $\pm$ 0.0040   &0.8525 $\pm$ 0.1477 \\
PHN-STCH & 2.8457 $\pm$ 0.0266 & 0.8044 $\pm$ 0.0595 & 2.1828 $\pm$ 0.0106 &0.8154 $\pm$ 0.1673& 2.7954 $\pm$ 0.0212 &0.4716 $\pm$ 0.1077 &0.7293 $\pm$ 0.0196   &0.7803 $\pm$ 0.1691 \\
PHN-EPO  & 2.8532 $\pm$ 0.0079 &0.9445 $\pm$ 0.1305 & 2.1669 $\pm$ 0.0296 & 0.8820 $\pm$ 0.1567 &2.7758 $\pm$ 0.0282 & 0.4960 $\pm$ 0.0884& 0.7049 $\pm$ 0.0042  &0.5362 $\pm$ 0.0509  \\
PHN-HVI             & 3.0023 $\pm$ 0.0003  &0.7339 $\pm$ 0.0488   & 2.3332 $\pm$ 0.0177  & 0.7823 $\pm$ 0.0591  & 2.9139 $\pm$ 0.0103  & 0.4620 $\pm$ 0.0384  &0.8309 $\pm$ 0.1717   &0.3679 $\pm$ 0.0565\\
\hline
A-SVH &\textbf{3.0942$\pm$  0.0106}  & \textbf{0.2225 $\pm$ 0.0134}& \textbf{2.4125 $\pm$ 0.0043}  &\textbf{0.2555 $\pm$ 0.0076} &\textbf{2.9505 $\pm$ 0.0047}  &\textbf{0.2262$\pm$ 0.0342} &\textbf{0.9676 $\pm$ 0.0125} &\textbf{0.3010$\pm$ 0.0498} \\

\bottomrule
\end{tabular}}

\caption{Multi-task learning performance comparison. Hypervolume (HV, higher is better) and $\Delta$-Spread ($\Delta$, lower is better) are reported for three image classification datasets (Multi-MNIST, Multi-Fashion, Fashion+MNIST) and one regression dataset (SARCOS). Our A-SVH-MOL is compared against five baseline PHN methods. Results are shown as mean $\pm$ standard deviation.}
\label{tab: performance_comparison}
\end{table}
\subsubsection*{Multi-Output Regression.}
The SARCOS dataset is a widely used benchmark for multi-task learning (MTL) and regression, particularly in robotics and control. We evaluate our model on a 7-task setup, where each task corresponds to predicting the torque of one of the seven robot arm joints. Each input is a 21-dimensional vector comprising joint positions (7), velocities (7), and accelerations (7). The output is a 7-dimensional vector of joint torques. The dataset contains $44,484$ training and $4,449$ test examples. We reserve $10\%$ of the training set for validation.

Our results, presented on the right in Table~\ref{tab: performance_comparison}, show that A-SVH-MOL achieves superior performance compared to all baseline methods. These results demonstrate that A-SVH-MOL performs remarkably well on high-dimensional multi-objective problems, achieving superior outcomes in both optimization quality and solution diversity. This indicates that our framework offers a promising approach for approximating the Pareto front in many-objective optimization tasks.
\subsection{Ablation Studies and Component Analysis}

\subsubsection{Impact of the Annealing Schedule} In this experiment, we analyze the impact of the annealing schedule on performance by comparing models with and without the annealing term. The results show that our proposed annealing strategy outperforms the cyclical annealing approach, confirming its effectiveness in maintaining training stability. Moreover, the findings validate our observation that integrating annealing into the Stein Variational Hypernetwork prevents the collapse issues commonly observed in vanilla or cyclical SVH models. 
\label{subsub:tradeoff}
\begin{table}[!ht]
\centering
\resizebox{\textwidth}{!}{
\begin{tabular}{ll||ccc||ccc}
\hline
\textbf{Category} & \textbf{Problem} & \textbf{SVH-LS} & \textbf{SVH-TCH} & \textbf{SVH-STCH} & \textbf{A-SVH-LS} & \textbf{A-SVH-TCH} & \textbf{A-SVH-STCH} \\ \hline
\multirow{11}{*}{\rotatebox{90}{\parbox{2.5cm}{\centering Real-world\\Problems}}} 
& RE21 &{\ul 0.8325 $\pm$ 0.0000}   &  0.8293 $\pm$ 0.0003 &0.8322 $\pm$ 0.0004   &0.8318 $\pm$ 0.0002   &  0.8306 $\pm$ 0.0003 & $\mathbf{0.8336 \pm 0.0004}$ \\
& RE22 &0.5188 $\pm$ 0.0357   & 0.4583 $\pm$ 0.0501 &0.4372 $\pm$ 0.0606   &0.5110 $\pm$ 0.0711  &0.5009 $\pm$ 0.0305  & 0.4498 $\pm$ 0.0421 \\
& RE23 &0.0653 $\pm$ 0.0304  & 0.4391 $\pm$ 0.2961  & 0.1012 $\pm$ 0.0607  &0.0626 $\pm$ 0.0325   & $\mathbf{0.8955 \pm 0.0662}$  &{\ul 0.7913 $\pm$ 0.4444}   \\
& RE24 & 1.1631 $\pm$ 0.0003  & 1.1525 $\pm$ 0.0009  &1.1545 $\pm$ 0.0003   &1.1635 $\pm$ 0.0000   & 1.1614 $\pm$ 0.0005  &$\mathbf{1.1636 \pm0.0004 }$  \\
& RE31 &1.3309 $\pm$ 0.0000   &$\mathbf{1.3310 \pm 0.0000}$   &1.3310 $\pm$ 0.0000   & 1.3309 $\pm$ 0.0000  &1.3309 $\pm$ 0.0000   & $\mathbf{1.3310 \pm 0.0000 }$ \\
& RE32 &1.3297 $\pm$ 0.0002   & 1.3298 $\pm$ 0.0001  &1.3297 $\pm$ 0.0001   & {\ul1.3299 $\pm$ 0.0001}  & 1.3298 $\pm$ 0.0001  & $\mathbf{1.3300 \pm 0.0003 }$ \\
& RE33 &0.9471 $\pm$ 0.0213   & {\ul1.0055 $\pm$ 0.0053}  &1.0127 $\pm$ 0.0026  & 0.9297 $\pm$ 0.0001   & 1.0022 $\pm$ 0.0027 & $\mathbf{1.0158 \pm 0.0007}$   \\
& RE34 &0.8430 $\pm$ 0.1248   & 0.9067 $\pm$ 0.0546 &0.9417 $\pm$ 0.0527   & 0.8432 $\pm$ 0.1245  & 0.9094 $\pm$ 0.0550  &0.9103 $\pm$ 0.0543   \\
& RE35 & 1.1945 $\pm$ 0.0746  &  1.2422 $\pm$ 0.0468 & {\ul 1.2458 $\pm$ 0.0643}  & 1.1940 $\pm$ 0.0752  &1.1815 $\pm$ 0.0511   & $\mathbf{1.2934 \pm 0.0008}$ \\
& RE36 &1.0183 $\pm$ 0.0002   & 1.0176 $\pm$ 0.0005  &1.0177 $\pm$ 0.0007   & $\mathbf{1.0184 \pm 0.0003} $  &1.0180 $\pm$ 0.0003   & 1.0180 $\pm$ 0.0002  \\
& RE37 &0.5938 $\pm$ 0.0000   &  0.8288 $\pm$ 0.0046 &0.8379 $\pm$ 0.0004   &0.6652 $\pm$ 0.0505   &  0.8261 $\pm$ 0.0024 & $\mathbf{0.8380 \pm 0.0008}$  \\ \hline
\multirow{3}{*}{\rotatebox{90}{\parbox{1.5cm}{\centering Multi-\\task}}} 
& Multi-MNIST &3.0803 $\pm$ 0.0033   & 3.0935 $\pm$ 0.0048  & 3.0877 $\pm$ 0.0049  & 3.0784 $\pm$ 0.0046  &3.0896 $\pm$ 0.0063   &\textbf{3.0942 $\pm$ 0.0106}  \\
& Multi-Fashion & 2.3946 $\pm$ 0.0074   & 2.4051 $\pm$ 0.0038  & 2.4001 $\pm$ 0.0034   & 2.3966 $\pm$ 0.0022  & 2.3944 $\pm$ 0.0021  & \textbf{2.4123 $\pm$ 0.0044}  \\
& Fashion+MNIST &2.8959 $\pm$ 0.0190  &2.9109 $\pm$ 0.0165   &  {\ul 2.9291 $\pm$ 0.0090} &2.8964 $\pm$ 0.0161   & 2.9242 $\pm$ 0.0127  & $\mathbf{2.9505 \pm 0.0047}$  \\ \hline
\end{tabular}
}
\caption{Comparison of hypervolume values: Vanilla SVH vs Annealed SVH. Best results in \textbf{bold}, second-best {\ul underlined}. Standard deviations shown where available (from multiple runs).}
\label{tab:vanilla_a_svh_comparison}
\end{table}

\textbf{A-SVH vs. Vanilla SVH} In Table~\ref{tab:vanilla_a_svh_comparison}, we report the hypervolume performance on both real-world problems and multi-task learning when comparing vanilla-SVH-MOL and A-SVH-MOL. These results indicate that our annealing schedule is better than vanilla-SVH-MOL without scheduling. Moreover, A-SVH-STCH exhibits a higher hypervolume in comparison with others. These figures are reasonable, as the strength of STCH scalarization was proved in the previous study \cite{lin2024smooth}.

\textbf{A-SVH vs. Cyclical Annealing}
In Table~\ref{tab:cyclical_svh}, we compare our annealed SVH-MOL with a baseline that employs a vanilla annealing schedule—cyclical annealing, where the annealing process is repeated with a fixed period. These results demonstrate the effectiveness and stability of our framework, as it performs better in almost all experiments. Furthermore, the use of an annealing strategy enhances the performance of SVH, with A-SVH consistently achieving higher results compared to SVH without the annealing strategy. 
\begin{table}[!t]
\centering
\resizebox{0.5\textwidth}{!}{
\begin{tabular}{lll}
\toprule
Dataset & Schedule & HV \\
\midrule
\multirow{2}{*}{Multi-MNIST} 
& Cyclical Annealing & $3.0758 \pm 0.0079$ \\
& Our Annealing & $\mathbf{3.0942 \pm 0.0106}$ \\

\midrule
\multirow{2}{*}{Multi-Fashion} 
& Cyclical Annealing & $2.3857 \pm 0.0098$ \\
& Our Annealing & $\mathbf{2.4123 \pm 0.0044}$ \\

\midrule
\multirow{2}{*}{Multi-Fashion+MNIST} 
& Cyclical Annealing & $2.9046 \pm 0.0036$ \\
& Our Annealing & $\mathbf{2.9505 \pm 0.0047}$ \\
\bottomrule
\end{tabular}
}
\caption{Performance comparison between our proposed annealing schedule (Eq. 12) and a standard cyclical annealing baseline on the multi-task image classification datasets. Our method achieves a higher and more stable Hypervolume (HV), demonstrating its superior design.}
\label{tab:cyclical_svh}
\end{table}

\begin{figure}[!t]
    \centering
    \includegraphics[width=0.9\linewidth]{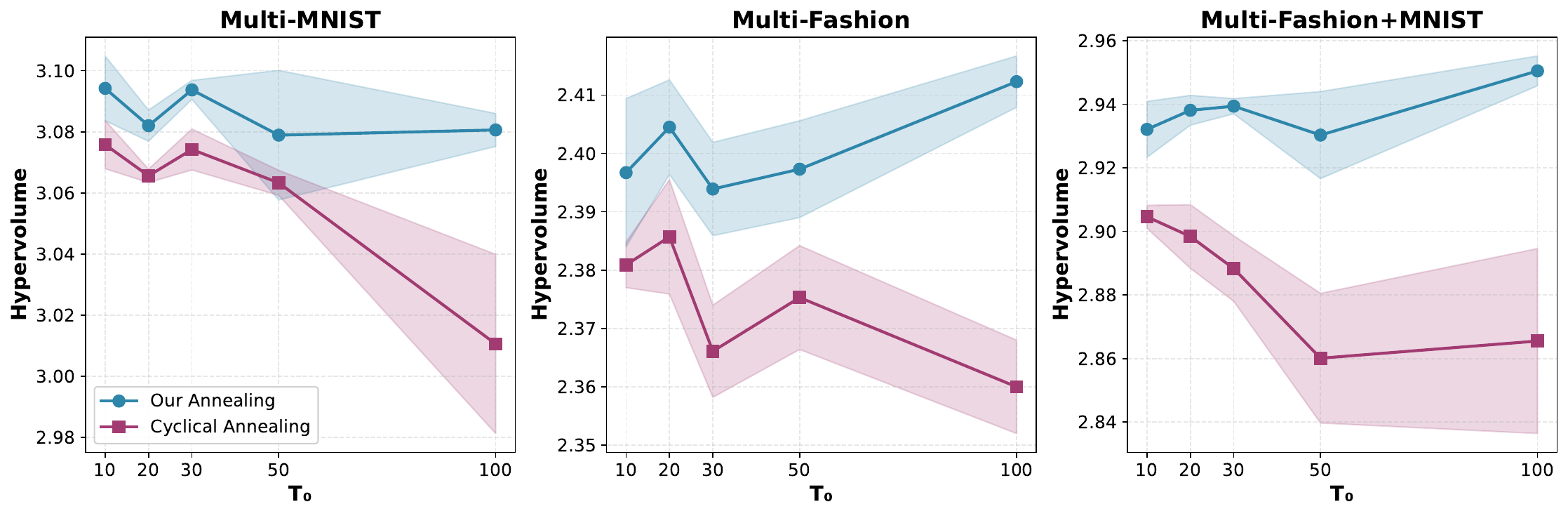}
    
\caption{Ablation study on the initial annealing period ($T_0$) on multi-task learning. The plots compare the Hypervolume (HV) performance of \textbf{(Left)} a standard cyclical annealing schedule against \textbf{(Right)} our proposed annealing method (Eq. 12). Our method shows stable and improving performance with a larger $T_0$, while the cyclical method becomes unstable and degrades.}
    \label{fig: trend_T_0}
\end{figure}

\textbf{Analysis of Annealing Period $T_0$} To systematically evaluate the effectiveness of the annealing mechanism in our proposed algorithm, we conduct ablation studies with varying initial warm-up periods of length $T_0$. This experimental design allows us to determine the optimal duration for the exploration phase and validate whether enhanced exploration during the initial epochs contributes to superior long-term convergence performance. We present the investigation in Figure~\ref {fig: trend_T_0}, including two approaches (cyclical annealing and our annealing). We experiment with $T_0$ varying numbers from 10 to 100 to explore the trend of the initial warm-up time on multi-task learning. These results indicate that our annealing maintains the stability across $T_0$, when it gains the lower standard deviation. Moreover, this trend shows that when training the model with a larger $T_0$, the HV results are higher. In contrast, the cyclical annealing reflects the inefficiency obviously. The trend is down, and the standard deviation is higher when $T_0$ increases. This difference demonstrates that our annealing SVH exhibits stability when the repulsive term is encouraged in the exploration process. This verifies our argument that exploring should be the first step in training. Pushing exploration at the last epoch is harmful, as illustrated in the cyclical annealing approach. 

\subsubsection{Convergence Performance Evaluation}
\begin{figure}[!ht]
    \centering
    \includegraphics[width=0.9\linewidth]{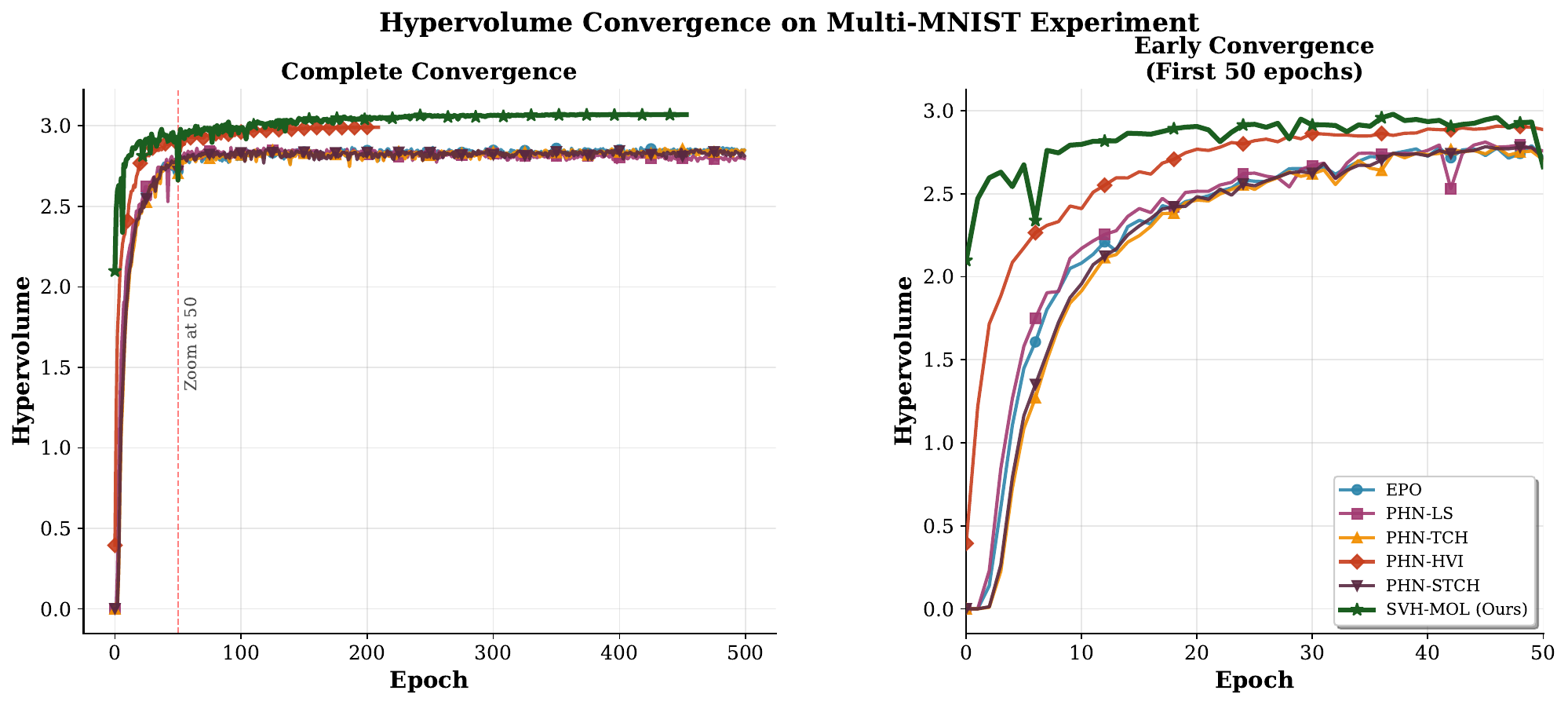}
    \caption{Convergence performance comparison on the Multi-MNIST dataset. \textbf{(Left)} Hypervolume (HV) over 500 training epochs. Our A-SVH-MOL (black line) converges faster and to a higher HV than all baseline methods. \textbf{(Right)} A zoomed-in view of the first 60 epochs, highlighting the initial fluctuations attributed to the annealing mechanism's exploration phase.}
    \label{fig: convergence}
\end{figure}

In Figure \ref{fig: convergence}, We conduct a comprehensive comparison of our proposed A-SVH-MOL method against several established baseline algorithms to evaluate their convergence characteristics on the Multi-MNIST dataset. The experimental results demonstrate the superior performance of A-SVH-MOL in terms of both convergence speed and final hypervolume achievement. As illustrated in the full convergence plot, A-SVH-MOL exhibits remarkable early-stage performance, rapidly achieving higher hypervolume values within the initial epochs compared to baseline methods. This rapid initial convergence indicates the effectiveness of our method in quickly identifying and exploiting promising regions of the Pareto frontier. 

The zoomed subplot focusing on the first 50 generations provides crucial insights into the early convergence behavior. While A-SVH-MOL demonstrates some fluctuation during this initial phase, we attribute this phenomenon to the beneficial effects of the annealing mechanism. These fluctuations represent the algorithm's adaptive exploration strategy, where the annealing process enables dynamic adjustment of the search intensity, allowing the method to escape local optima and discover better solutions more effectively

\subsection{Hyperparameter Sensitivity Analysis}

\subsubsection{Effect of Diversity Parameter $\alpha$}
Table~\ref{tab:ablation_alpha} shows the sensitivity of the diversity parameter $\alpha$ on multi-task learning. Results indicate that large $\alpha$ values ($\alpha > 0.1$) increase diversity but reduce convergence, as seen by higher $\Delta$-Spread and lower HV. 
In contrast, moderate values ($10^{-2} \leq \alpha \leq 10^{-5}$) achieve a better balance, yielding more stable and higher HV scores with moderate diversity. 
This suggests that small $\alpha$ promotes stable convergence, whereas large $\alpha$ favors diverse but less convergent solutions.
\begin{table}[!t]
\centering
\resizebox{0.95\columnwidth}{!}{%
\begin{tabular}{lcccccc}
\hline
 & \multicolumn{2}{c}{\textbf{Multi-MNIST}} & \multicolumn{2}{c}{\textbf{Multi-Fashion}} & \multicolumn{2}{c}{\textbf{Multi-Fashion+MNIST}} \\
\cline{2-3} \cline{4-5} \cline{6-7}
\textbf{$\alpha$} & \textbf{HV} & \textbf{$\Delta$-Spread}  & \textbf{HV} & \textbf{$\Delta$-Spread}  & \textbf{HV} & \textbf{$\Delta$-Spread}  \\
\hline
$1$ & $3.0687 \pm 0.0041$ & $0.2217 \pm 0.0076$ & $1.1111 \pm 0.9409$ & $0.4918 \pm 0.2909$ & $2.8359 \pm 0.0328$ & $\mathbf{0.2259 \pm 0.0503}$  \\
\hline
$0.5$ & $2.9909 \pm 0.0649$ & $0.2476 \pm 0.0298$ & $2.3675 \pm 0.0274$ & $\mathbf{0.2166 \pm 0.0098}$ & $2.8549 \pm 0.0118$ & $0.2702 \pm 0.0406$ \\
\hline
$0.1$ & $2.8154 \pm 0.0697$ & $\mathbf{0.1820 \pm 0.0149}$ & $2.3805 \pm 0.0138$ & $0.2416 \pm 0.0400$ & $2.8154 \pm 0.0697$ & $0.3169 \pm 0.0474$ \\
\hline
$1\times e^{-2}$ & $\mathbf{3.0942 \pm 0.0106}$ & $0.2225 \pm 0.0134$ & $2.4077 \pm 0.0107$ & $0.2536 \pm 0.0211$ & $2.8660 \pm 0.0630$ & $0.2985 \pm 0.0880$ \\
\hline
$1\times e^{-3}$ & $3.0878 \pm 0.0080$ & $0.2267 \pm 0.0130$ & $\mathbf{2.4125 \pm 0.0043}$ & $0.2555 \pm 0.0076$ & $2.9110 \pm 0.0058$ & $0.2614 \pm 0.0032$ \\
\hline
$1\times e^{-4}$ & $3.0904 \pm 0.0023$ & $0.2188 \pm 0.0146$ & $2.3983 \pm 0.0019$ & $0.2887 \pm 0.0152$ & $\mathbf{2.9505 \pm 0.0047}$ & $0.2262 \pm 0.0342$ \\
\hline
$1\times e^{-5}$ & $3.0895 \pm 0.0064$ & $0.2194 \pm 0.0115$ & $2.3919 \pm 0.0039$ & $0.2758 \pm 0.0387$ & $2.9267 \pm 0.0171$ & $0.2527 \pm 0.0032$ \\
\hline
\end{tabular}
}
\caption{Hyperparameter sensitivity analysis for the diversity parameter $\alpha$ on multi-task learning datasets. Both Hypervolume (HV, higher is better) and $\Delta$-Spread ($\Delta$, lower is better) are reported.}
\label{tab:ablation_alpha}
\end{table}
\begin{figure}[!t]
\centering
\begin{minipage}[c]{0.3\textwidth}
    \centering
    \small
    \begin{tabular}{lc}
    \hline
    \textbf{Bandwidth} $h$ & \textbf{HV} \\
    \hline
    \textbf{Median Heuristic} & \textbf{3.0942} $\pm$ \textbf{0.0106}  \\ \hline
    $h = 1e^{-6}$ &3.0839 $\pm$ 0.0011   \\
    $h = 1e^{-5}$ & 3.0902$\pm$ 0.0017  \\
    $h = 1e^{-4}$ & 3.0926 $\pm$ 0.0030 \\
    $h = 1e^{-3}$ & 3.0289$\pm$ 0.0041  \\
    $h = 1e^{-2}$ & 2.9275$\pm$ 0.0040  \\
    $h = 1e^{-1}$ & 2.9879 $\pm$ 0.0509   \\
    $h = 1e^{0}$ & 3.0717$\pm$  0.0141  \\
    \hline
    \end{tabular}
\end{minipage}\hfill
\begin{minipage}[c]{0.6\textwidth}
    \centering
    \includegraphics[width=\textwidth]{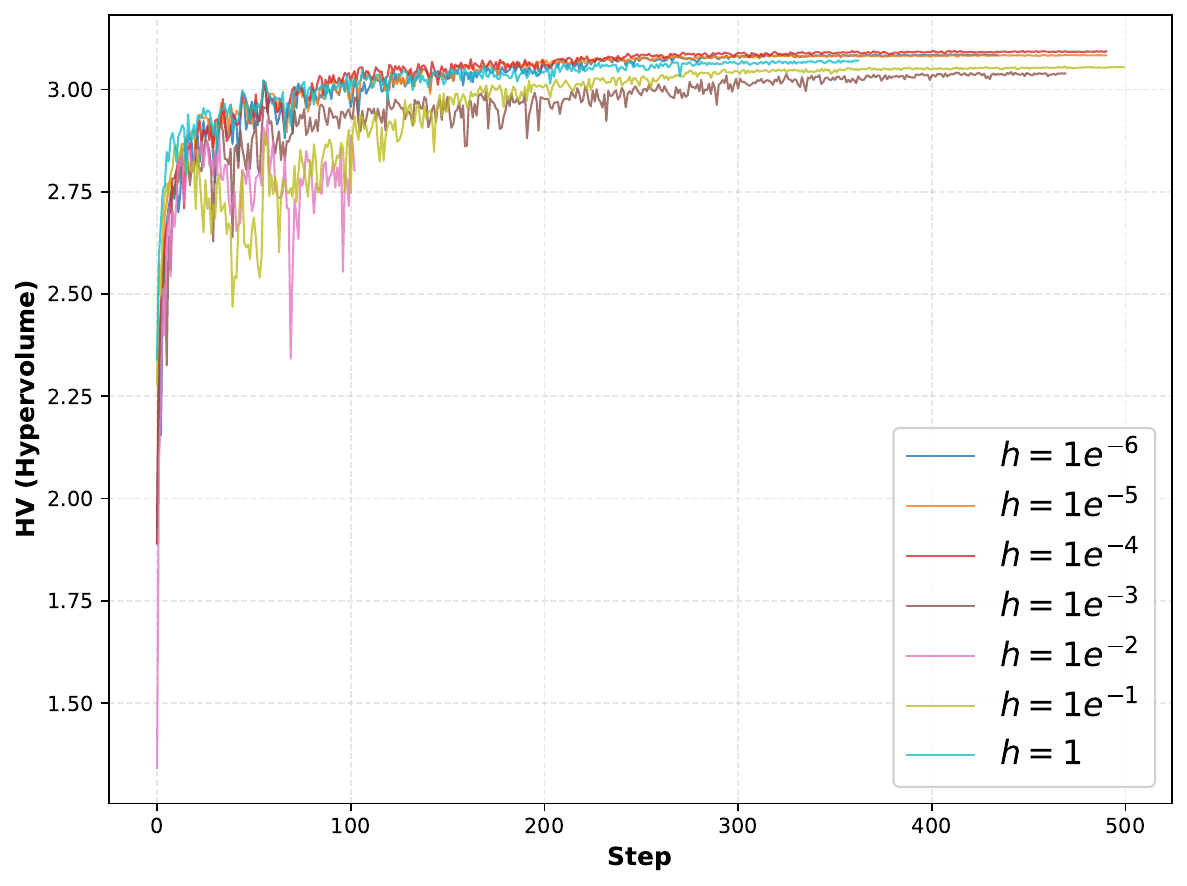}
\end{minipage}
\caption{Effect of kernel bandwidth ($h$) on Multi-MNIST performance. \textbf{(Left)} Table of final Hypervolume (HV) values, comparing various fixed bandwidths to the adaptive median heuristic. \textbf{(Right)} Convergence curves for each bandwidth setting. The median heuristic (black line) provides the most robust and highest-performing result.}
\label{fig:bandwidth_ablation}
\end{figure}
\subsubsection{Effect of Kernel Bandwidth $h$}

Figure~\ref{fig:bandwidth_ablation} presents a comprehensive analysis of kernel bandwidth selection, comparing fixed values with the adaptive median heuristic approach. The bandwidth parameter $h$ critically influences particle interactions during training, governing the trade-off between local exploitation and global exploration. 

As shown in the table, the median heuristic bandwidth achieves the best hypervolume performance ($3.0942 \pm 0.0106$). Analysis of fixed bandwidth values reveals a clear trend: very small bandwidths ($h = 1e^{-6}$ to $h = 1e^{-4}$) maintain competitive performance close to the median heuristic, with $h = 1e^{-5}$ achieving $3.0902 \pm 0.0017$. However, as bandwidth increases beyond $h = 1e^{-3}$, performance begins to degrade, with $h = 1e^{-2}$ dropping to $2.9275 \pm 0.0040$ and $h = 1e^{-1}$ further declining to $2.9879 \pm 0.0509$. Interestingly, $h = 1e^{0}$ shows a partial recovery to $3.0717 \pm 0.0141$, though still underperforming the median heuristic. These results demonstrate that while small fixed bandwidths can achieve reasonable performance, the median heuristic provides superior and more robust results by adaptively adjusting to the particle distribution, avoiding the need for manual tuning and ensuring consistent convergence. The convergence curves (right panel) provide additional insights, showing that while some configurations achieve rapid initial progress, all experience optimization instability during early training phases. These results underscore the importance of adaptive bandwidth selection, as the median heuristic automatically adjusts to the particle distribution, avoiding the drawback of manually tuned fixed values.
\subsubsection{Effect of Number of Particles}
Table \ref{tab:particle_count} presents the results of varying the number of particles in the training process and its influence on the hypervolume (HV) performance. As the number of particles increases, the computational burden of both the kernel function evaluation and the gradient update step grows substantially, leading to longer training times per epoch. Furthermore, we observe a decreasing trend in the hypervolume as the number of particles increases. In particular, while a moderate number of particles can enhance exploration and diversity, an excessively large set appears to dilute the model’s optimization effectiveness, resulting in lower HV values. These findings suggest that simply increasing the sample set does not necessarily yield better performance. Instead, there exists an optimal balance between computational efficiency and optimization quality. Notably, our SVH-PSL model achieves its best performance with a relatively small number of particles, demonstrating that it can efficiently approximate the Pareto front without relying on a large sample set.
\label{subsec:particle_ablation}

\begin{table}[!t]
\centering

\begin{tabular}{lcc}
\hline
Particles ($n$) & HV  & Time/Epoch (s) \\
\hline
4 & \textbf{3.0942$\pm$  0.0106}  & 64.49\\
8 & 3.0877 $\pm$  0.0051 &   145.30\\
16 & 3.0808 $\pm$  0.0051 & 447.23 \\
32 & 2.9950  $\pm$ 0.0037& 1603.27\\
\hline
\end{tabular}
\caption{Ablation study on the number of particles ($n$) used during training on the Multi-MNIST dataset. The table shows the trade-off between performance (Hypervolume, HV) and computational cost (Time/Epoch in seconds).}
\label{tab:particle_count}
\end{table}
\section{Limitations and Future Work}
\label{sec:limitations}

While SVH-MOL demonstrates strong performance across various multi-objective problems, several limitations warrant discussion:

\subsection*{Computational Complexity}
The pairwise kernel computations in SVGD scale quadratically with the number of particles ($O(n^2)$), which can become prohibitive for large particle counts. While our hypernetwork approach mitigates this through parameter sharing, very large-scale applications may require approximate kernel methods or sampling strategies.

\subsection*{Hyperparameter Sensitivity}
Our method requires careful tuning of several hyperparameters, particularly the diversity parameter $\alpha$ and annealing schedule parameters ($T_0$, $\tau$). Although we provide extensive sensitivity analysis, automatic hyperparameter optimization would enhance practical usability.

\subsection*{High-Dimensional Objective Spaces}
While we demonstrate effectiveness on problems with up to 7 objectives, the "curse of dimensionality" may affect performance in many-objective optimization (problems with >10 objectives). The repulsive force mechanism may become less effective in very high-dimensional spaces.

\subsection*{Theoretical Guarantees}
Although empirically effective, theoretical convergence guarantees for our annealed SVGD approach in multi-objective settings remain an open question. Formal analysis of convergence properties would strengthen the methodological foundation.

\subsection*{Future Directions}
Although our approach is effective in finding the convergence and diversity Pareto set, a fixed kernel selection is not universal. We suggest that exploring an adaptive kernel might be more suitable in future work. This adaptive kernel eliminates a set of hyperparameters, using learnable parameters instead in kernel calculation. Besides reducing computational costs, this approach automatically optimizes the trade-off between the driving and repulsive terms. Following our SVH-MOL framework, the kernel matrix is defined in the objective space, where an effective adaptive kernel can capture the geometric structure of the objective dimensions and promote moving to the optimal region better. 

\section{Conclusion}
In this paper, we introduce SVH-MOL, a novel framework that integrates Stein Variational Gradient Descent (SVGD) with hypernetwork-based techniques to efficiently approximate the entire Pareto set in multi-objective optimization. By incorporating an annealing schedule, our approach effectively balances convergence and diversity, addressing a key challenge in multi-objective learning. Extensive experiments on synthetic benchmarks and real-world multi-task learning problems demonstrate that SVH-MOL outperforms existing methods. These findings highlight the robustness and scalability of our method, paving the way for future advancements in controllable and efficient optimization techniques.


\section*{Acknowledgment}
This research was funded by Vingroup Innovation Foundation (VINIF) under project code VinIF.2024.DA113. 

\bibliographystyle{ACM-Reference-Format}
\bibliography{acmart.bib}

@inproceedings{zhou2024beyond,
  title={Beyond one-preference-fits-all alignment: Multi-objective direct preference optimization},
  author={Zhou, Zhanhui and Liu, Jie and Shao, Jing and Yue, Xiangyu and Yang, Chao and Ouyang, Wanli and Qiao, Yu},
  booktitle={Findings of the Association for Computational Linguistics: ACL 2024},
  pages={10586--10613},
  year={2024}
}

@article{wu2023fine,
  title={Fine-grained human feedback gives better rewards for language model training},
  author={Wu, Zeqiu and Hu, Yushi and Shi, Weijia and Dziri, Nouha and Suhr, Alane and Ammanabrolu, Prithviraj and Smith, Noah A and Ostendorf, Mari and Hajishirzi, Hannaneh},
  journal={Advances in Neural Information Processing Systems},
  volume={36},
  pages={59008--59033},
  year={2023}
}

@inproceedings{liu2024mftcoder,
  title={Mftcoder: Boosting code llms with multitask fine-tuning},
  author={Liu, Bingchang and Chen, Chaoyu and Gong, Zi and Liao, Cong and Wang, Huan and Lei, Zhichao and Liang, Ming and Chen, Dajun and Shen, Min and Zhou, Hailian and others},
  booktitle={Proceedings of the 30th ACM SIGKDD Conference on Knowledge Discovery and Data Mining},
  pages={5430--5441},
  year={2024}
}

@inproceedings{liu2025pareto,
  title={Pareto set learning for multi-objective reinforcement learning},
  author={Liu, Erlong and Wu, Yu-Chang and Huang, Xiaobin and Gao, Chengrui and Wang, Ren-Jian and Xue, Ke and Qian, Chao},
  booktitle={Proceedings of the AAAI Conference on Artificial Intelligence},
  volume={39},
  number={18},
  pages={18789--18797},
  year={2025}
}

@article{haishan2024preference,
  title={Preference-optimized pareto set learning for blackbox optimization},
  author={Haishan, Zhang and Das, Diptesh and Tsuda, Koji},
  journal={arXiv preprint arXiv:2408.09976},
  year={2024}
}

@article{zhang2023hypervolume,
  title={Hypervolume maximization: A geometric view of pareto set learning},
  author={Zhang, Xiaoyuan and Lin, Xi and Xue, Bo and Chen, Yifan and Zhang, Qingfu},
  journal={Advances in Neural Information Processing Systems},
  volume={36},
  pages={38902--38929},
  year={2023}
}

@article{deb2002fast,
  title={A fast and elitist multiobjective genetic algorithm: NSGA-II},
  author={Deb, Kalyanmoy and Pratap, Amrit and Agarwal, Sameer and Meyarivan, TAMT},
  journal={IEEE transactions on evolutionary computation},
  volume={6},
  number={2},
  pages={182--197},
  year={2002},
  publisher={Ieee}
}

@article{10.1162/neco.1997.9.8.1735,
author = {Hochreiter, Sepp and Schmidhuber, J\"{u}rgen},
title = {Long Short-Term Memory},
year = {1997},
issue_date = {November 15, 1997},
publisher = {MIT Press},
address = {Cambridge, MA, USA},
volume = {9},
number = {8},
issn = {0899-7667},
url = {https://doi.org/10.1162/neco.1997.9.8.1735},
doi = {10.1162/neco.1997.9.8.1735},
journal = {Neural Comput.},
month = nov,
pages = {1735–1780},
numpages = {46}
}

@article{duncan2023geometry,
  title={On the geometry of Stein variational gradient descent},
  author={Duncan, Andrew and N{\"u}sken, Nikolas and Szpruch, Lukasz},
  journal={Journal of Machine Learning Research},
  volume={24},
  number={56},
  pages={1--39},
  year={2023}
}

@inproceedings{ba2021understanding,
  title={Understanding the variance collapse of SVGD in high dimensions},
  author={Ba, Jimmy and Erdogdu, Murat A and Ghassemi, Marzyeh and Sun, Shengyang and Suzuki, Taiji and Wu, Denny and Zhang, Tianzong},
  booktitle={International Conference on Learning Representations},
  year={2021}
}

@inproceedings{zhuo2018message,
  title={Message passing Stein variational gradient descent},
  author={Zhuo, Jingwei and Liu, Chang and Shi, Jiaxin and Zhu, Jun and Chen, Ning and Zhang, Bo},
  booktitle={International Conference on Machine Learning},
  pages={6018--6027},
  year={2018},
  organization={PMLR}
}

@incollection{YANG2014197,
title = {Chapter 14 - Multi-Objective Optimization},
editor = {Xin-She Yang},
booktitle = {Nature-Inspired Optimization Algorithms},
publisher = {Elsevier},
address = {Oxford},
pages = {197-211},
year = {2014},
isbn = {978-0-12-416743-8},
doi = {https://doi.org/10.1016/B978-0-12-416743-8.00014-2},
url = {https://www.sciencedirect.com/science/article/pii/B9780124167438000142},
author = {Xin-She Yang}
}

@article{steuer1983interactive,
  title={An interactive weighted Tchebycheff procedure for multiple objective programming},
  author={Steuer, Ralph E and Choo, Eng-Ung},
  journal={Mathematical programming},
  volume={26},
  pages={326--344},
  year={1983},
  publisher={Springer}
}

@article{deb2013evolutionary,
  title={An evolutionary many-objective optimization algorithm using reference-point-based nondominated sorting approach, part I: solving problems with box constraints},
  author={Deb, Kalyanmoy and Jain, Himanshu},
  journal={IEEE transactions on evolutionary computation},
  volume={18},
  number={4},
  pages={577--601},
  year={2013},
  publisher={IEEE}
}

@inproceedings{schaffer2014multiple,
  title={Multiple objective optimization with vector evaluated genetic algorithms},
  author={Schaffer, J David},
  booktitle={Proceedings of the first international conference on genetic algorithms and their applications},
  pages={93--100},
  year={2014},
  organization={Psychology Press}
}

@article{tanabe2020easy,
  title={An easy-to-use real-world multi-objective optimization problem suite},
  author={Tanabe, Ryoji and Ishibuchi, Hisao},
  journal={Applied Soft Computing},
  volume={89},
  pages={106078},
  year={2020},
  publisher={Elsevier}
}

@article{10.1162/106365600568202,
    author = {Zitzler, Eckart and Deb, Kalyanmoy and Thiele, Lothar},
    title = {Comparison of Multiobjective Evolutionary Algorithms: Empirical Results},
    journal = {Evolutionary Computation},
    volume = {8},
    number = {2},
    pages = {173-195},
    year = {2000},
    month = {06},
    issn = {1063-6560},
    doi = {10.1162/106365600568202},
    url = {https://doi.org/10.1162/106365600568202},
    eprint = {https://direct.mit.edu/evco/article-pdf/8/2/173/1493199/106365600568202.pdf},
}

@ARTICLE{zitzler1999multiobjective,
  title={Multiobjective evolutionary algorithms: a comparative case study and the strength Pareto approach},
  author={Zitzler, Eckart and Thiele, Lothar},
  journal={IEEE transactions on Evolutionary Computation},
  volume={3},
  number={4},
  pages={257--271},
  year={1999},
  publisher={IEEE}
}

@article{d2021annealed,
  title={Annealed stein variational gradient descent},
  author={D'Angelo, Francesco and Fortuin, Vincent},
  journal={arXiv preprint arXiv:2101.09815},
  year={2021}
}

@article{liu2021profiling,
  title={Profiling pareto front with multi-objective stein variational gradient descent},
  author={Liu, Xingchao and Tong, Xin and Liu, Qiang},
  journal={Advances in neural information processing systems},
  volume={34},
  pages={14721--14733},
  year={2021}
}

@article{Nguyen_Dinh_Nguyen_Hoang_Le_2025, title={Improving Pareto Set Learning for Expensive Multi-objective Optimization via Stein Variational Hypernetworks}, volume={39}, url={https://ojs.aaai.org/index.php/AAAI/article/view/34167}, DOI={10.1609/aaai.v39i18.34167}, abstractNote={Expensive multi-objective optimization problems (EMOPs) are common in real-world scenarios where evaluating objective functions is costly and involves extensive computations or physical experiments. Current Pareto set learning methods for such problems often rely on surrogate models like Gaussian processes to approximate the objective functions. These surrogate models can become fragmented, resulting in numerous small uncertain regions between explored solutions. When using acquisition functions such as the Lower Confidence Bound (LCB), these uncertain regions can turn into pseudo-local optima, complicating the search for globally optimal solutions. To address these challenges, we propose a novel approach called SVH-PSL, which integrates Stein Variational Gradient Descent (SVGD) with Hypernetworks for efficient Pareto set learning. Our method addresses the issues of fragmented surrogate models and pseudo-local optima by collectively moving particles in a manner that smooths out the solution space. The particles interact with each other through a kernel function, which helps maintain diversity and encourages the exploration of underexplored regions. This kernel-based interaction prevents particles from clustering around pseudo-local optima and promotes convergence towards globally optimal solutions. Our approach aims to establish robust relationships between trade-off reference vectors and their corresponding true Pareto solutions, overcoming the limitations of existing methods. Through extensive experiments across both synthetic and real-world MOO benchmarks, we demonstrate that SVH-PSL significantly improves the quality of the learned Pareto set, offering a promising solution for expensive multi-objective optimization problems.}, number={18}, journal={Proceedings of the AAAI Conference on Artificial Intelligence}, author={Nguyen, Minh-Duc and Dinh, Phuong Mai and Nguyen, Quang-Huy and Hoang, Long P. and Le, Dung D.}, year={2025}, month={Apr.}, pages={19677-19685} }

@article{ha2016hypernetworks,
  title={Hypernetworks},
  author={Ha, David and Dai, Andrew and Le, Quoc V},
  journal={arXiv preprint arXiv:1609.09106},
  year={2016}
}

@misc{vijayakumar2000sarcos,
  author = {Vijayakumar, S.},
  title = {The {SARCOS} Dataset},
  year = {2000},
  url = {http://www.gaussianprocess.org/gpml/data},
  note = {Accessed: 2023-03-09}
}

@article{ai2023stein,
  title={Stein variational gradient descent with multiple kernels},
  author={Ai, Qingzhong and Liu, Shiyu and He, Lirong and Xu, Zenglin},
  journal={Cognitive Computation},
  volume={15},
  number={2},
  pages={672--682},
  year={2023},
  publisher={Springer}
}

@article{liu2016stein,
  title={Stein variational gradient descent: A general purpose bayesian inference algorithm},
  author={Liu, Qiang and Wang, Dilin},
  journal={Advances in neural information processing systems},
  volume={29},
  year={2016}
}

@article{navon2020learning,
  title={Learning the pareto front with hypernetworks},
  author={Navon, Aviv and Shamsian, Aviv and Chechik, Gal and Fetaya, Ethan},
  journal={arXiv preprint arXiv:2010.04104},
  year={2020}
}

@article{lin2019pareto,
  title={Pareto multi-task learning},
  author={Lin, Xi and Zhen, Hui-Ling and Li, Zhenhua and Zhang, Qing-Fu and Kwong, Sam},
  journal={Advances in neural information processing systems},
  volume={32},
  year={2019}
}

@inproceedings{mahapatra2020multi,
  title={Multi-task learning with user preferences: Gradient descent with controlled ascent in pareto optimization},
  author={Mahapatra, Debabrata and Rajan, Vaibhav},
  booktitle={International Conference on Machine Learning},
  pages={6597--6607},
  year={2020},
  organization={PMLR}
}

@article{yu2020gradient,
  title={Gradient surgery for multi-task learning},
  author={Yu, Tianhe and Kumar, Saurabh and Gupta, Abhishek and Levine, Sergey and Hausman, Karol and Finn, Chelsea},
  journal={Advances in Neural Information Processing Systems},
  volume={33},
  pages={5824--5836},
  year={2020}
}

@inproceedings{chen2018gradnorm,
  title={Gradnorm: Gradient normalization for adaptive loss balancing in deep multitask networks},
  author={Chen, Zhao and Badrinarayanan, Vijay and Lee, Chen-Yu and Rabinovich, Andrew},
  booktitle={International conference on machine learning},
  pages={794--803},
  year={2018},
  organization={PMLR}
}

@article{lin2024smooth,
  title={Smooth Tchebycheff Scalarization for Multi-Objective Optimization},
  author={Lin, Xi and Zhang, Xiaoyuan and Yang, Zhiyuan and Liu, Fei and Wang, Zhenkun and Zhang, Qingfu},
  journal={arXiv preprint arXiv:2402.19078},
  year={2024}
}

@inproceedings{kendall2018multi,
  title={Multi-task learning using uncertainty to weigh losses for scene geometry and semantics},
  author={Kendall, Alex and Gal, Yarin and Cipolla, Roberto},
  booktitle={Proceedings of the IEEE conference on computer vision and pattern recognition},
  pages={7482--7491},
  year={2018}
}

@inproceedings{liu2019end,
  title={End-to-end multi-task learning with attention},
  author={Liu, Shikun and Johns, Edward and Davison, Andrew J},
  booktitle={Proceedings of the IEEE/CVF conference on computer vision and pattern recognition},
  pages={1871--1880},
  year={2019}
}

@article{chen2025gradient,
  title={Gradient-Based Multi-Objective Deep Learning: Algorithms, Theories, Applications, and Beyond},
  author={Chen, Weiyu and Zhang, Xiaoyuan and Lin, Baijiong and Lin, Xi and Zhao, Han and Zhang, Qingfu and Kwok, James T},
  journal={arXiv preprint arXiv:2501.10945},
  year={2025}
}

@ARTICLE{4358754,
  author={Zhang, Qingfu and Li, Hui},
  journal={IEEE Transactions on Evolutionary Computation}, 
  title={MOEA/D: A Multiobjective Evolutionary Algorithm Based on Decomposition}, 
  year={2007},
  volume={11},
  number={6},
  pages={712-731},
  doi={10.1109/TEVC.2007.892759}}

@inproceedings{nguyen2024high,
  title={High-Dimensional Bayesian Optimization via Random Projection of Manifold Subspaces},
  author={Nguyen, Quoc-Anh Hoang and Tran, The Hung},
  booktitle={Joint European Conference on Machine Learning and Knowledge Discovery in Databases},
  pages={288--305},
  year={2024},
  organization={Springer}
}

@article{lin2022pareto,
  title={Pareto set learning for expensive multi-objective optimization},
  author={Lin, Xi and Yang, Zhiyuan and Zhang, Xiaoyuan and Zhang, Qingfu},
  journal={Advances in neural information processing systems},
  volume={35},
  pages={19231--19247},
  year={2022}
}

@article{crawshaw2020multi,
  title={Multi-task learning with deep neural networks: A survey},
  author={Crawshaw, Michael},
  journal={arXiv preprint arXiv:2009.09796},
  year={2020}
}

@article{sener2018multi,
  title={Multi-task learning as multi-objective optimization},
  author={Sener, Ozan and Koltun, Vladlen},
  journal={Advances in neural information processing systems},
  volume={31},
  year={2018}
}

@article{milojkovic2019multi,
  title={Multi-gradient descent for multi-objective recommender systems},
  author={Milojkovic, Nikola and Antognini, Diego and Bergamin, Giancarlo and Faltings, Boi and Musat, Claudiu},
  journal={arXiv preprint arXiv:2001.00846},
  year={2019}
}

@article{jannach2022multi,
  title={Multi-objective recommender systems: Survey and challenges},
  author={Jannach, Dietmar},
  journal={arXiv preprint arXiv:2210.10309},
  year={2022}
}

@article{zaizi2023multi,
  title={Multi-objective optimization with recommender systems: A systematic review},
  author={Zaizi, Fatima Ezzahra and Qassimi, Sara and Rakrak, Said},
  journal={Information Systems},
  volume={117},
  pages={102233},
  year={2023},
  publisher={Elsevier}
}

@article{Craft2006,
  author = {Craft, DL and Halabi, TF and Shih, HA and Bortfeld, TR},
  title = {Approximating Convex Pareto Surfaces in Multiobjective Radiotherapy Planning},
  journal = {Medical Physics},
  volume = {33},
  number = {9},
  pages = {3399--3407},
  year = {2006},
  month = {September},
  doi = {10.1118/1.2335486},
  pmid = {17022236}
}

@article{Marler2004,
  author = {Marler, R. T. and Arora, J. S.},
  title = {Survey of Multi-Objective Optimization Methods for Engineering},
  journal = {Structural and Multidisciplinary Optimization},
  volume = {26},
  number = {6},
  pages = {369--395},
  year = {2004},
  month = {April},
  doi = {10.1007/s00158-003-0368-6}
}

@misc{deist2021multiobjectivelearningpredictpareto,
      title={Multi-Objective Learning to Predict Pareto Fronts Using Hypervolume Maximization}, 
      author={Timo M. Deist and Monika Grewal and Frank J. W. M. Dankers and Tanja Alderliesten and Peter A. N. Bosman},
      year={2021},
      eprint={2102.04523},
      archivePrefix={arXiv},
      primaryClass={cs.LG},
      url={https://arxiv.org/abs/2102.04523}, 
}

@inproceedings{Hoang2023,
author = {Hoang, Long P. and Le, Dung D. and Tuan, Tran Anh and Thang, Tran Ngoc},
title = {Improving pareto front learning via multi-sample hypernetworks},
year = {2023},
isbn = {978-1-57735-880-0},
publisher = {AAAI Press},
url = {https://doi.org/10.1609/aaai.v37i7.25953},
doi = {10.1609/aaai.v37i7.25953},
booktitle = {Proceedings of the Thirty-Seventh AAAI Conference on Artificial Intelligence and Thirty-Fifth Conference on Innovative Applications of Artificial Intelligence and Thirteenth Symposium on Educational Advances in Artificial Intelligence},
articleno = {884},
numpages = {9},
series = {AAAI'23/IAAI'23/EAAI'23}
}

@article{tuan2024framework,
  title={A framework for controllable pareto front learning with completed scalarization functions and its applications},
  author={Tuan, Tran Anh and Hoang, Long P and Le, Dung D and Thang, Tran Ngoc},
  journal={Neural Networks},
  volume={169},
  pages={257--273},
  year={2024},
  publisher={Elsevier}
}

\appendix
\end{document}